%% file: 0_main.tex
\documentclass{article}

\usepackage{microtype}
\usepackage{enumitem}
\usepackage{multirow}
\usepackage{amsfonts}
\usepackage{subfigure}
 \usepackage{amsmath}
\usepackage{subcaption}
\usepackage{comment}
\usepackage{xcolor}
\usepackage{xspace}
\usepackage{amsmath,bm}
\usepackage{graphicx}
\usepackage{listings}
\usepackage{booktabs} 

\usepackage{hyperref}

\newcommand{\sysname}{\textsc{Hippocampus}\xspace}

\usepackage[accepted]{mlsys2025}

\mlsystitlerunning{}

\begin{document}

\twocolumn[
\mlsystitle{\sysname: An Efficient and Scalable Memory Module for Agentic AI}



\mlsyssetsymbol{equal}{*}

\begin{mlsysauthorlist}
\mlsysauthor{Yi Li}{to,goo}
\mlsysauthor{Lianjie Cao}{goo}
\mlsysauthor{Faraz Ahmed}{goo}
\mlsysauthor{Puneet Sharma}{goo}
\mlsysauthor{Bingzhe Li }{to}
\end{mlsysauthorlist}

\mlsysaffiliation{to}{The University of Texas at Dallas}
\mlsysaffiliation{goo}{Hewlett Packard Enterprise (HPE) Labs}

\mlsyscorrespondingauthor{Yi Li}{yi.li3@utdallas.edu}

\mlsyskeywords{Machine Learning, MLSys}

\vskip 0.3in

\begin{abstract}
Agentic AI require persistent memory to store user-specific histories beyond the limited context window of LLMs. Existing memory systems use dense vector databases or knowledge-graph traversal (or hybrid), incurring high retrieval latency and poor storage scalability. We introduce \sysname, an agentic memory management system that uses compact binary signatures for semantic search and lossless token-ID streams for exact content reconstruction. Its core is a Dynamic Wavelet Matrix (DWM) that compresses and co-indexes both streams to support ultra-fast search in the compressed domain, thus avoiding costly dense-vector or graph computations. This design scales linearly with memory size, making it suitable for long-horizon agentic deployments. Empirically, our evaluation shows that \sysname reduces end-to-end retrieval latency by up to 31$\times$ and cuts per-query token footprint by up to 14$\times$, while maintaining accuracy on both LoCoMo and LongMemEval benchmarks.

\end{abstract}
]



\printAffiliationsAndNotice{Work done while the first author was a research intern at Hewlett Packard Enterprise (HPE) Network and Distributed Systems Lab (NDSL).}  

\input{1_intro_v1.1}

\input{2_back-motiv_v1.1}

\section{Design of \sysname}\label{sec: design}

We present the technical design of \sysname, a system built for scalable, high-throughput agentic AI memory management. At the core of the design is a dual-representation strategy that simultaneously supports exact, high-fidelity content retrieval and fast, approximate semantic search. Central to this strategy is the Dynamic Wavelet Matrix (DWM)—a compressed, bit-level data structure that we develop and employ to index both representations. This approach enables \sysname to achieve high data density while maintaining low-latency query performance. We begin by outlining the high-level system architecture (Section~\ref{sec: overall}), which illustrates the data flow during memory construction and retrieval.

A theoretical analysis of the DWM’s efficiency and scalability is provided in Appendix~\ref{app: dwm_time_space}, along with a quantitative comparison showing that our query scheme (Section~\ref{sec: RI_HB}) maintains a small accuracy gap relative to dense vector retrieval (Appendix~\ref{app: acc_gap}).

\subsection{Overall Architecture}\label{sec: overall}
The architecture of \sysname memory module is composed of two primary components: a pipeline for memory construction (as shown in Figure~\ref{fig:overall_construction}) and a pipeline for memory querying (as shown in Figure~\ref{fig:overall_query}).

\textbf{Memory Construction Pipeline.} As shown in Figure~\ref{fig:overall_construction}, the memory construction pipeline begins by ingesting raw data, such as a dialogue turn in LoCoMo~\cite{maharana2024evaluating} (see Section~\ref{sec: setting} for details), which is then processed through two parallel steps. 
First, content serialization converts the unstructured text into a canonical sequence of tokens. 
Concurrently, metadata extraction captures essential contextual information. 
For the LoCoMo dataset, this includes the speaker/role, a high-resolution timestamp, and the start ($\alpha$) and end ($\beta$) indices of each utterance within the serialized token list.

The core of \sysname is a dual-representation strategy, realized through two distinct Dynamic Wavelet Matrices (DWMs) (see Section~\ref{sec: dwm} for details): a Content DWM for exact data representation and a Signature DWM for efficient, approximate semantic search. To construct the Content DWM, each token in the serialized sequence (i.e., from content serialization) is mapped to its corresponding integer token ID, which is then converted into its binary representation. These binary codes are vertically arranged to form a bit matrix, constituting the Content DWM. In parallel, the Random Indexing \& Token Signature module computes a low-dimensional binary hash—or signature—for each token (see Section~\ref{sec: RI_HB} for details). This process produces a compact token signature sequence, which is used to construct the Signature DWM in the same manner. This dual-matrix structure enables \sysname to support both precise content retrieval and fast, semantics-based similarity queries within a unified framework.

\textbf{Memory Retrieval Process.} We now describe the query process, illustrated in Figure~\ref{fig:overall_query}, which leverages the two constructed DWMs (as shown in Figure~\ref{fig:overall_construction}) to enable highly efficient agentic AI memory retrieval. The query pipeline begins when a natural language query is received. First, the query is processed by a lightweight LLM Prompt module to extract a set of salient keywords. These keywords are then passed through the same Random Indexing \& Token Signature module used during memory construction, converting them into their corresponding binary signatures. These query signatures are used to perform a fast, approximate search—e.g., based on Hamming distance (see Section~\ref{sec: RI_HB} for details)—against the Signature DWM. This initial pass rapidly filters the entire memory space and identifies a small set of candidate data segments by retrieving their associated metadata blocks. The StartIndex ($\alpha$) and EndIndex ($\beta$) from each candidate’s metadata serve as direct pointers for exact, indexed retrieval from the Content DWM. This step reconstructs the original, full-resolution token sequences for the candidate segments. The retrieved content and its corresponding metadata are then returned as the final Retrieved Result.
This two-stage design allows \sysname to efficiently search over vast conversational histories by using the compact Signature DWM as a fast, low-cost index into the high-fidelity Content DWM.

With the end-to-end data flow established, we now dive into the core technical components that underpin the \sysname architecture.
The following subsections are organized as follows: we first provide a detailed formulation of the Dynamic Wavelet Matrix (DWM) (Section~\ref{sec: dwm}), which serves as the fundamental data structure in \sysname. We then describe the Random Indexing and Hamming Ball search mechanism (Section~\ref{sec: RI_HB}), which is used to generate robust token signatures and perform approximate search.

\subsection{Dynamic Wavelet Matrix}\label{sec: dwm}
The core data structure underlying both the content store and the signature store in \sysname is the Dynamic Wavelet Matrix (DWM). The DWM is a novel data structure we develop specifically to support efficient and incremental indexing for agentic AI memory. It serves as an append-friendly adaptation of the conventional static Wavelet Matrix (WM)~\cite{gog2014optimized,claude2012wavelet}, a well-established structure in information retrieval for compressing and indexing large sequences. By extending the WM to support dynamic updates while preserving its compression and query efficiency, the DWM enables high-throughput memory construction and retrieval in continuously evolving agentic systems.

A key limitation of the conventional wavelet matrix is its static nature—it is designed to be built once over a fixed collection~\cite{resnikoff2012wavelet}. This design is fundamentally incompatible with agentic memory workload, which is modeled as a high-throughput, append-only stream. 
Rebuilding a static WM for every new dialogue turn (see Section~\ref{sec: background} for details) would be computationally prohibitive.

\textbf{Notation and Structure.} We conceptually represent DWM as a bit-matrix over an integer sequence $S[0, \cdots, n-1]$.
Given a dictionary (as illustrated in Figure~\ref{fig:token_id}) of size $\sigma$~\cite{rajaraman2024toward}, each unique token can be encoded with $\lceil \log_2 \sigma \rceil$ bits.
Accordingly, DWM consists of $l$ independent bit-vectors, denoted as $\bm{B}^0, \cdots, \bm{B}^{l-1}$, each of the length $n$.
This ever-growing (as shown in Figure~\ref{fig:DWM_Construction}) matrix of size $\lceil \log_2 \sigma \rceil \times n$, where $\lceil \log_2 \sigma \rceil$ is fixed and depends solely on dictionary size. It is constructed such that each $\bm{B}^k[i]$ stores the $k$-th bit of the symbol $S[i]$.
This bit-matrix representation is highly compressible and efficiently supports fundamental sequence operations ($access(i)$, $rank(c, i)$, and $select(c, k)$) (defined in Section~\ref{sec: background}).

\subsubsection{Dynamic Construction}
\begin{figure}[!t]
\centering
  \includegraphics[width=0.9\columnwidth]{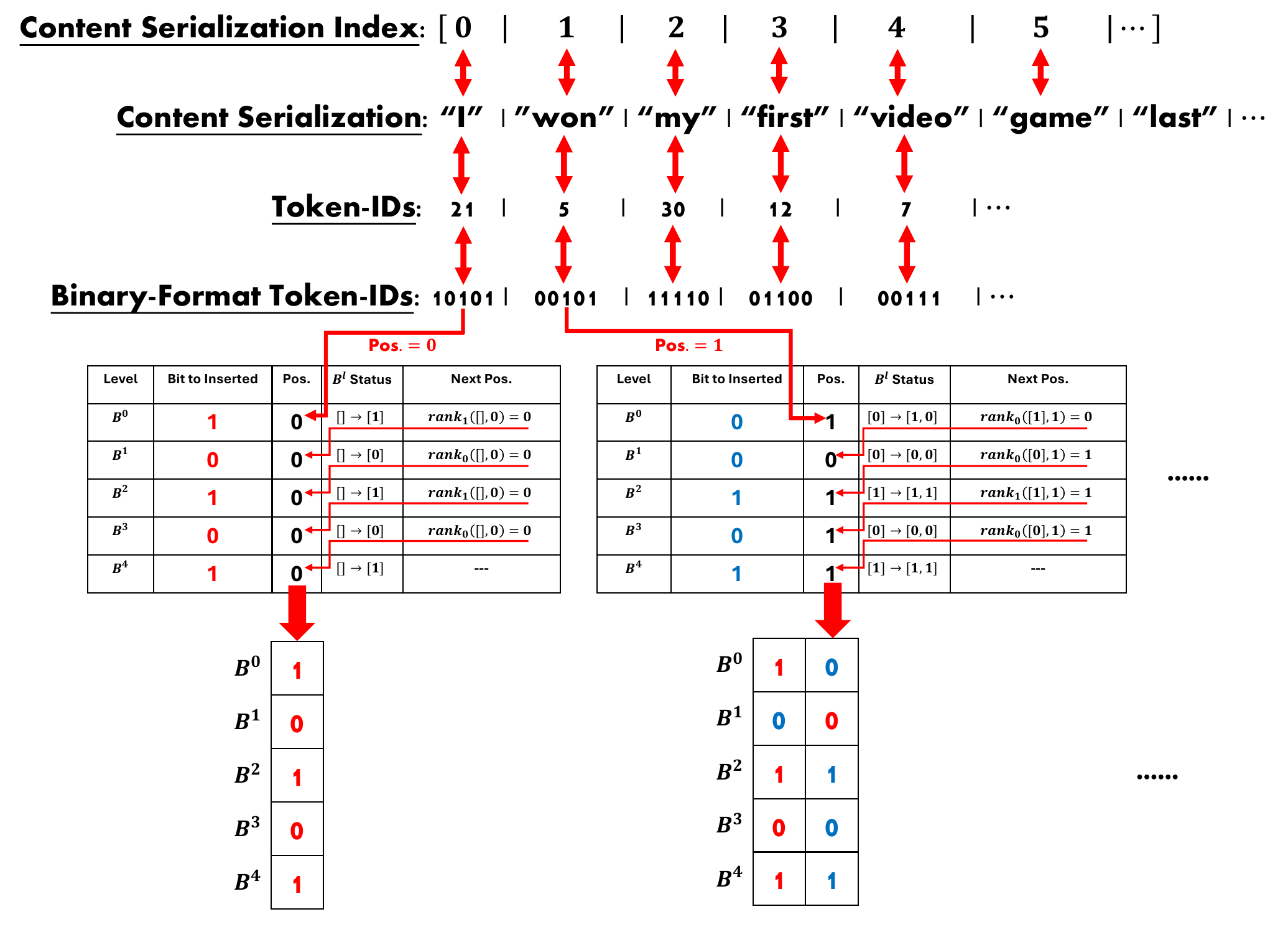}
  \caption{Construction process of Dynamic Wavelet Matrix.}
  \label{fig:DWM_Construction}
  \vspace{-12pt}
\end{figure}
We construct the DWM through a sequence of $append(s)$ operations, each adding a new symbol $s$ (an integer in $[0, \sigma)$) to the end of sequence $S$. Figure~\ref{fig:DWM_Construction} illustrates this process. To append symbol $s$ at the new global position $Pos = n$ (initially $n=0$ for an empty sequence), we perform a single top-down traversal of the $l$ levels:
\begin{enumerate}
[topsep=0pt,itemsep=-1ex,partopsep=1ex,parsep=1ex, leftmargin=*]
    \item Write the most significant bit of $s$ to $\bm{B}^{0}[Pos]$. For example, in Figure~\ref{fig:DWM_Construction}, if $s$ has binary representation $(10101)_2$, we append $1$ as the new bit in $B^0$.
    \item Determine the local position for $s$ in the next level. If the bit just appended was 0, the symbol will go into the “zero” side of the next level; if it was 1, into the “one” side. We compute this by counting the number of 0s or 1s before $Pos$. For instance, if we appended a 1 in step 1, we set a temporary index $p = rank_1(\bm{B}^0, Pos)$, which gives the count of 1s in $\bm{B}^0$ up to (but not including) the new position. This $p$ will be the position of the symbol in the next level’s bit-vector.
    \item Move to the next level and append the second bit of $s$ at position $p$ of $\bm{B}^{1}$. If this bit is 0, then for level 2 the position resets to $p = rank_0(\bm{B}^1, p)$ (counting how many 0s precede the just-appended bit in $\bm{B}^1$). If the bit is 1, then $p$ becomes $Z_1 + rank_1(\bm{B}^1, p)$, where $Z_1$ is the total number of 0s in $B^1$ (i.e., the starting offset of the “one” portion at level 1). This gives the insertion position for level 2.
    \item Repeat this process for all $l$ bits of $s$, descending through the levels. By the end, $s$ is fully inserted in the DWM, and $n$ increases by 1.
\end{enumerate}

Through this construction process, the DWM is incrementally maintained in a strictly append-only and efficient manner.  
Each $append$ operation runs in \( \mathcal{O}(l) \) time—equivalently \( \mathcal{O}(\log \sigma) \), which is typically much smaller than \( n \)—assuming constant-time support for rank operations ($rank_{0/1}$) on the bit-vectors.

Our DWM design provides exactly the query primitives needed for \sysname.  
The Content DWM supports direct access to any token via the $access(i)$ operation.  
The Signature DWM enables efficient approximate membership queries using $rank$ and $select$ operations, which we leverage to perform Hamming-distance-based searches for relevant signatures.  
We next describe how these queries operate within the DWM and how they enable memory recall in \sysname.

\subsubsection{Memory Recall with Dynamic Wavelet Matrix}
\begin{algorithm}[t]
\caption{DWM \textsc{rank}$(c,i)$ operation}
\label{alg:dwm-rank}
\small
\begin{algorithmic}[1]
\REQUIRE Token signature $\bm{c}$ with bits $(b_0,\ldots,b_{l-1})$ where $b_0$ is MSB; global prefix length $i$; Sign. DWM $\bm{D}$ with level bit-vectors $\bm{D}.B[0\ldots l-1]$ and zero-counts $\bm{D}.Z[0\ldots l-1]$
\ENSURE $rank(c,i)$ = \#occurrences of $c$ in $S[0,i)$
\STATE $(b_0,b_1,\ldots,b_{l-1}) \leftarrow \text{BitsMSBFirst}(\bm{c})$
\STATE $p_L \leftarrow 0$;\quad $p_R \leftarrow i$
\FOR{$k=0$ to $l-1$}
  \IF{$b_k = 0$}
    \STATE $p_L \leftarrow rank_0(\bm{D}.B[k], p_L)$
    \STATE $p_R \leftarrow rank_0(\bm{D}.B[k], p_R)$
  \ELSE
    \STATE $Z_k \leftarrow \bm{D}.Z[k]$
    \STATE $p_L \leftarrow Z_k + rank_1(\bm{D}.B[k], p_L)$
    \STATE $p_R \leftarrow Z_k + rank_1(\bm{D}.B[k], p_R)$
  \ENDIF
\ENDFOR
\STATE \textbf{return} $p_R - p_L$
\end{algorithmic}
\end{algorithm}
\begin{algorithm}[t]
\caption{DWM $select(c,j)$ operation}
\label{alg:dwm-select}
\small
\begin{algorithmic}[1]
\REQUIRE Token signature $\bm{c}$ with bits $(b_0,\ldots,b_{l-1})$ where $b_0$ is MSB; 1-based occurrence $j$; sequence length $n$; Sign.\ DWM $\bm{D}$ with level bit-vectors $\bm{D}.B[0\ldots l-1]$ and zero-counts $\bm{D}.Z[0\ldots l-1]$
\ENSURE $select(c,j)$ = global position of the $j$-th $c$ in $S$
\STATE $(b_0,b_1,\ldots,b_{l-1}) \leftarrow \text{BitsMSBFirst}(\bm{c})$
\STATE $p_L \leftarrow 0$;\quad $p_R \leftarrow n$
\FOR{$k=0$ to $l-1$}
  \IF{$b_k = 0$}
    \STATE $p_L \leftarrow rank_0(\bm{D}.B[k], p_L)$
    \STATE $p_R \leftarrow rank_0(\bm{D}.B[k], p_R)$
  \ELSE
    \STATE $Z_k \leftarrow \bm{D}.Z[k]$
    \STATE $p_L \leftarrow Z_k + rank_1(\bm{D}.B[k],p_L)$
    \STATE $p_R \leftarrow Z_k + rank_1(\bm{D}.B[k],p_R)$
  \ENDIF
\ENDFOR
\STATE $occ \leftarrow p_R - p_L$
\IF{$j > occ$}
  \STATE \textbf{return} $\textbf{NULL}$ 
\ENDIF
\STATE $p \leftarrow p_L + (j-1)$
\FOR{$k=l-1$ to $0$ \textbf{step} $-1$}
  \IF{$b_k = 0$}
    \STATE $p \leftarrow select_0(\bm{D}.B[k],p+1)$
  \ELSE
    \STATE $Z_k \leftarrow \bm{D}.Z[k]$
    \STATE $p \leftarrow select_1(\bm{D}.B[k],(p - Z_k)+1)$
  \ENDIF
\ENDFOR
\STATE \textbf{return} $p$
\end{algorithmic}
\end{algorithm}
\vspace{-0.1in}

When a query is issued, \sysname uses the Signature DWM to identify likely relevant memory indices, and then uses the Content DWM to reconstruct the content at those indices.  
This process relies on the DWM’s ability to efficiently count and locate symbols.  
In the Signature DWM, each “symbol” is a compact binary signature representing a token.  
A natural language query is transformed into a set of such signature symbols \( \{ \bm{c}_1, \bm{c}_2, \ldots, \bm{c}_m \} \) via the random indexing step.  
Our goal is to find memory entries where all (or many) of these query signatures co-occur. 
We accomplish this by using the DWM $rank$ and $select$ primitives to traverse the Signature DWM efficiently.

\textbf{Searching in Signature DWM.}  
Suppose we have a particular signature $\bm{c}$ (a binary code of length $l$ bits) and we want to quickly find all positions in the Signature DWM where $\bm{c}$ appears.  
We can use $rank(\bm{c}, i)$ to count occurrences of $\bm{c}$ up to any position $i$, and $select(\bm{c}, j)$ to retrieve the position of the $j$-th occurrence.  
Algorithm~\ref{alg:dwm-rank} outlines the $rank$ query.  
Starting from the most significant bit of $\bm{c}$, we use the bit values to narrow an interval $[p_L, p_R)$ as we descend the levels.  
Initially, $p_L = 0$ and $p_R = i$ (meaning we consider the prefix $S[0..i-1]$).  
At each level $k$, if the $k$-th bit of $\bm{c}$ is 0, we map the current interval to the zero-prefixed subarray of the next level by setting  
$p_L \leftarrow rank_0(\bm{B}^k, p_L)$ and $p_R \leftarrow rank_0(\bm{B}^k, p_R)$.  
If the bit is 1, we map to the one-prefixed subarray by setting  
$p_L \leftarrow Z_k + rank_1(\bm{B}^k, p_L)$ and $p_R \leftarrow Z_k + rank_1(\bm{B}^k, p_R)$,  
where $Z_k$ is the total number of 0s in $\bm{B}^k$.  
After processing all $l$ bits, the length of the final interval $(p_R - p_L)$ equals the number of occurrences of $\bm{c}$ in $S[0..i-1]$.

To retrieve the actual positions of occurrences, we use the $select(\bm{c}, j)$ operation, outlined in Algorithm~\ref{alg:dwm-select}.  
We first find the total number of occurrences $occ = rank(\bm{c}, n)$ in the entire sequence of length $n$.  
If $j > occ$, the $j$-th occurrence does not exist.  
Otherwise, we know the $j$-th occurrence lies in the interval $[p_L, p_R)$ obtained by running the $rank$ procedure (Algorithm~\ref{alg:dwm-rank}) to the end of the sequence ($i = n$).  
We set $p = p_L + (j - 1)$, which is the index of this occurrence in the bottom level.  
We then lift this index back up through the levels.  
For each level $k$ (going from $l-1$ up to $0$):  
if $b_k = 0$ (the $k$-th bit of $\bm{c}$ is 0), we call $p \leftarrow select_0(\mathbf{D}.B[k], p + 1)$,  
which finds the global position of the $(p + 1)$-th 0-bit in level $k$.  
If $b_k = 1$, we set $p \leftarrow select_1(\mathbf{D}.B[k], (p - Z_k) + 1)$,  
which finds the global position of the $(p - Z_k + 1)$-th 1-bit in level $k$ (accounting for the offset of the one-block).  
After lifting through all levels, $p$ gives the global position in $S$ of the $j$-th occurrence of $\bm{c}$.

In \sysname, we use these primitives to execute memory queries as follows.  
Given a set of query signatures: $\{\bm{c}_1, \ldots, \bm{c}_m\}$ extracted from the user’s query,  
we first identify the least frequent signature $\bm{c}_{\min}$ by comparing $rank(\bm{c}_i, n)$ for all $i$.  
We then iterate through each occurrence of $\bm{c}_{\min}$ in the Signature DWM.  
For the $j$-th occurrence (where $j$ ranges from 1 to $occ = rank(\bm{c}_{\min}, n)$),  
we find its global position $i = select(\bm{c}_{\min}, j)$.  
This position $i$ corresponds to a specific token in the memory sequence $S$.  
We retrieve the metadata entry whose range $[\alpha, \beta]$ covers $i$  
(recall that each memory entry’s start and end indices are stored in its metadata).  
This metadata tells us the span of token indices for that memory entry.  
If the query contains multiple keywords, we can quickly verify whether the other query signatures $\bm{c}_{2 \ldots m}$ appear in the same span  
by checking if $rank(\bm{c}_k, \beta) - rank(\bm{c}_k, \alpha) > 0$ for each $k$.  
Entries that pass this check are collected as candidate results.

\begin{algorithm}[!t]
\caption{DWM \textsc{access}$(i)$ operation}
\label{alg:dwm-access}
\small
\begin{algorithmic}[1]
\REQUIRE Global position $i$; Content DWM $\bm{D}$ with level bit-vectors $\bm{D}.B[0\ldots l-1]$ and zero-counts $\bm{D}.Z[0\ldots l-1]$
\ENSURE \textsc{access}$(i)$ = the symbol $S[i]$
\STATE $p \leftarrow i$
\STATE $\textit{bits} \leftarrow [\ ]$
\FOR{$k=0$ to $l-1$}
  \STATE $b \leftarrow \bm{D}.B[k][p]$
  \STATE $\textit{bits}.\texttt{append}(b)$
  \IF{$b = 0$}
    \STATE $p \leftarrow rank_0(\bm{D}.B[k],\, p)$
  \ELSE
    \STATE $Z_k \leftarrow \bm{D}.Z[k]$
    \STATE $p \leftarrow Z_k + rank_1(\bm{D}.B[k],\, p)$
  \ENDIF
\ENDFOR
\STATE \textbf{return} \text{SymbolFromBits}$(\textit{bits})$
\end{algorithmic}
\end{algorithm}


\textbf{Retrieving from Content DWM.}  
Finally, for each candidate memory entry identified via the above process, we perform a lossless reconstruction of its content using the Content DWM.  
This is achieved through the $access(i)$ primitive applied over the range $[\alpha, \beta]$ of token positions.  
Algorithm~\ref{alg:dwm-access} shows how a single symbol is retrieved by $access(i)$.  
We start at the top level with the global position $i$.  
At level $0$, we read the bit $b_0 = \bm{B}^0[i]$, which is the most significant bit of the symbol at $S[i]$.  
We append $b_0$ to a bit buffer and then determine the position at the next level:  
if $b_0 = 0$, we set $i_1 = rank_0(\bm{B}^0, i)$ (the number of 0s up to position $i$ in level 0);  
if $b_0 = 1$, we set $i_1 = Z_0 + rank_1(\bm{B}^0, i)$ (the number of 0s in level 0 plus the number of 1s up to $i$).  
We then move to level 1, read $b_1 = \bm{B}^1[i_1]$, append it, and update the position for level 2 in a similar fashion.

After we descend through all $l$ levels, we have collected bits $(b_0, b_1, \ldots, b_{l-1})$,  
which constitute the binary representation of $S[i]$.  
We then convert these bits back to the original token-id (an integer) using $\text{SymbolFromBits}$.  
In practice, we execute $access(i)$ for each position $i$ in the range $[\alpha, \beta]$  
to retrieve the entire sequence of token-ids for that memory entry, and then detokenize to reconstruct the text.

\subsection{Random Indexing and Hamming Ball}\label{sec: RI_HB}

While the DWM supports efficient keyword-exact matching, many queries require semantic-level retrieval for improved accuracy and robustness.  
To enable this capability, \sysname converts each token into a compact, context-aware binary signature.  
Instead of using static embeddings or precomputed vectors, we adopt a lightweight streaming random indexing mechanism~\cite{indyk1998approximate} that continuously integrates local contextual information during memory construction.

Specifically, let $D$ (e.g., 1024) denote the embedding dimensionality.  
At initialization, each token $v$ is assigned a sparse random base vector $\bm{r}_v = \{ -1, 0, +1 \}^D$ with exactly $t$ non-zero entries placed randomly (half $+1$ and half $-1$).  
These vectors remain fixed throughout content serialization.  
As the conversation stream arrives, we maintain a sliding window $W(i)$ around each token $S[i]$ and aggregate its contextual embedding via $\bm{e}_{i} = \sum_{j \in W(i)} \bm{r}_{S[j]}$, ensuring that tokens appear in slightly different semantic states depending on their conversational context~\cite{kanerva2000random}.  
After one streaming pass, each token has a fully contextualized embedding $\bm{e}_i$.  
Directly hashing all $D$ dimensions would incur unnecessary cost,  
so \sysname selects only the $d$ ($d \ll D$) most activated components:  
$\mathcal{I}_{i} = \text{Top-}d(\vert \bm{e}_{i} \vert)$.  
A binary signature is then formed:
$$
\bm{s}_{i}[k]=
\begin{cases}
    1, \quad \bm{e}_{i}[\mathcal{I}_{i}[k]] > 0\\
    0, \quad \bm{e}_{i}[\mathcal{I}_{i}[k]] \leq 0
\end{cases}
\quad k = 1,\cdots,d
$$

During querying (Figure~\ref{fig:overall_query}), an LLM extracts a small set of keywords from the natural language query that best describe the user’s intent.  
Each keyword is then converted into a $d$-bit signature using the same streaming random setting used during memory construction.  
We then perform an efficient Hamming-ball search on the Signature DWM.  
For each keyword signature $\bm{s}_q$ and a stored signature $\bm{s}_i$, we first compute a bitwise XOR, which returns a $d$-bit mask where 1s indicate mismatched bit positions.  
We then apply \texttt{POPCOUNT}~\cite{Sun2016RevisitingPO}, a native CPU instruction that counts the number of 1s in the mask in constant time, thus directly yielding the Hamming distance $\text{HammingDist}(\bm{s}_q, \bm{s}_i)$.  
A candidate is preserved only if this distance does not exceed a small threshold $r$, meaning we search within a Hamming-ball defined as:  
$\{ \bm{s}_i \mid \text{HammingDist}(\bm{s}_q, \bm{s}_i) \leq r \}$, 
so that only entries differing in at most $r$ bits (out of the $d$ bits) are considered semantically relevant and passed forward for subsequent metadata validation.

\section{Evaluation}\label{sec: evaluation}
\begin{table*}[]
\centering
\caption{Overall comparison of different memory modules across four tasks in LoCoMo benchmark: 
\textbf{Single-Hop}, \textbf{Multi-Hop}, \textbf{Temporal}, and \textbf{Open-Domain}.
We report the default metrics (F1 and BLEU-1) together with an LLM-as-a-Judge score reflecting human-aligned evaluation of answer quality. Reported F1 and BLEU-1 are multiplied by 100 for easier comparison and visualization.
}
\label{tab:overall_locomo_accuracy}
\resizebox{\linewidth}{!}{%
\begin{tabular}{l|ccc|ccc|ccc|ccc}
\toprule[1.5pt]
\multirow{2}{*}{Memory Module} & \multicolumn{3}{c|}{Single-Hop}  & \multicolumn{3}{c|}{Multi-Hop}   & \multicolumn{3}{c|}{Temporal}    & \multicolumn{3}{c}{Open-Domain} \\ \cline{2-13} 
                        & F1    & BLEU-1 & LLM-as-a-Judge & F1    & BLEU-1 & LLM-as-a-Judge & F1    & BLEU-1 & LLM-as-a-Judge & F1    & BLEU-1 & LLM-as-a-Judge \\ \midrule[1pt]
ReadAgent             & 8.78  & 5.93   & 1.03               & 5.44  & 5.03   & 1.01               & 11.24 & 11.12  & 1.08               & 9.32  & 8.1    & 1.45               \\
MemoryBank            & 5.05  & 3.97   & 2.00               & 6.02  & 5.89   & 1.12               & 9.85  & 9.92   & 1.03               & 7.9   & 7.97   & 2.09               \\
MemGPT               & 25.43 & 17.68  & 1.91               & 9.11  & 8.82   & 1.06               & 26.48 & 26.19  & 1.02               & 39.74 & 40.03  & 1.92               \\
A-mem                  & 19.82 & 19.86  &  2.66              & 12.97 & 12.81  &  1.85              & 34.63 & 34.87  &  2.18              & 41    & 41.41  &  2.72              \\
MemoryOS              & 32.5  & 30.13  & 2.76               & 28.61 & 26.81  & 1.79               & 25.08 & 25.08  & 2.61               & 41.51 & 41.43  & 2.59               \\
MemOS                  & \textbf{39.24} & \textbf{40.76}  &  2.75              & 30.11 & 30.91  & 2.56               & 31.06 & 31.34  & 2.81               & 40.31 & 40.51  & 2.60               \\ \midrule[1pt]
\sysname            & 34.36 & 30.04  &     \textbf{3.08}           & \textbf{31.97} & \textbf{31.85}  &  \textbf{3.22}              & \textbf{38.3}  & \textbf{37.35}  &    \textbf{2.94}            & \textbf{48.38} & \textbf{46.8}   &  \textbf{2.97}              \\ \bottomrule[1.5pt]
\end{tabular}%
}
\end{table*}

\subsection{Experimental Setup}\label{sec: setting}
\textbf{Dataset.}  
We adopt two of the most recent and widely used benchmarks designed to assess the long-term contextual memory capabilities of agentic AI:  
LoCoMo~\cite{maharana2024evaluating} and LongMemEval~\cite{wu2024longmemeval}.  
For a detailed description of the datasets, please refer to Appendix~\ref{app: experimental_setting}.

\textbf{Metric.}  
For LoCoMo, we adopt its default automatic evaluation metrics: F1~\cite{opitz2019macro} and BLEU-1~\cite{yang2008extending},  
which measure lexical overlap and token-level correctness in question answering tasks.  
For LongMemEval, the benchmark uses accuracy as its principal metric,  
defined as the fraction of evaluation questions answered correctly~\cite{wu2024longmemeval}.  
Beyond these standard metrics, we also introduce a \textit{LLM-as-a-Judge} score~\cite{gu2024survey}  
to better capture semantic correctness and deeper reasoning quality (Range: $[1, 2, 3, 4, 5]$).  
Refer to Appendix~\ref{app: prompt} for details.
In addition to accuracy-oriented metrics, we evaluate efficiency along two axes:
\begin{itemize}[topsep=0pt, itemsep=-1ex, partopsep=1ex, parsep=1ex, leftmargin=*]
    \item \textbf{Avg. Token Consumption:} average number of tokens read and processed per query, reflecting memory retrieval cost;
    \item \textbf{Avg. Total:} mean time from the moment memory recall is triggered to the moment the retrieved context is delivered (used for constructing the final prompt).  
    Within this, we further decompose and report \textbf{Avg. Search}, which captures the pure retrieval cost.
\end{itemize}

\textbf{Software and Hardware.}  
All experiments were conducted on a HPE DL380a Gen11 server with 2$\times$ Intel Xeon Platinum 8470 CPU, 4 $\times$ NVIDIA H100 GPUs, and 1~TB of DDR4 DRAM.
The software environment includes Ubuntu 22.04.5 LTS, Python 3.10.12, PyTorch 2.7.0, and CUDA 12.9 for GPU acceleration.

\vspace{-0.1in}
\subsection{Overall Comparison}\label{sec: overall_comparison}
\begin{figure}[!t]
\centering
  \includegraphics[width=0.95\columnwidth]{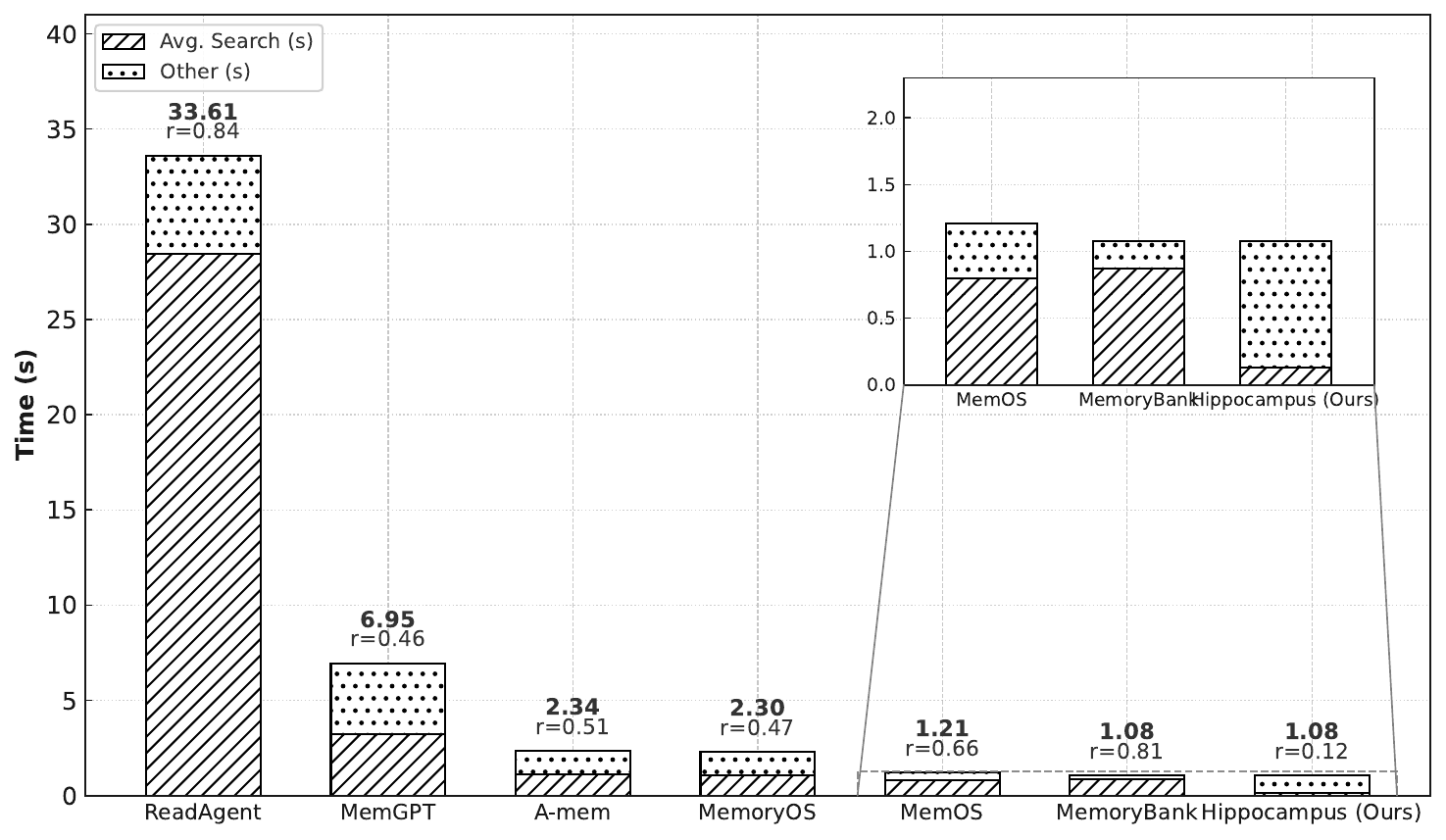}
  \caption{Average query retrieval latency (seconds) for various memory systems on LoCoMo dataset.}
  \label{fig:locomo_search_ratio}
  \vspace{-12pt}
\end{figure}
\begin{figure}[!t]
\centering
  \includegraphics[width=0.95\columnwidth]{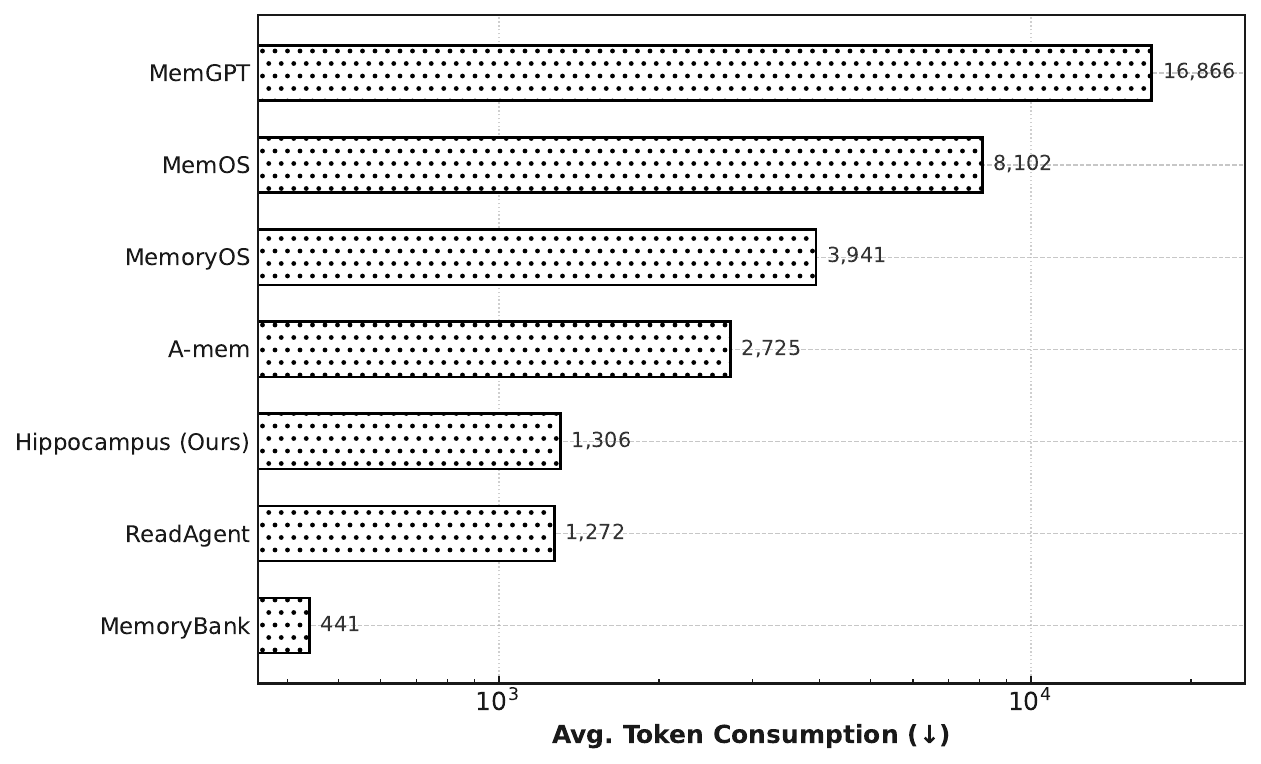}
  \caption{Average number of tokens consumption per query by each memory system on LoCoMo dataset.}
  \label{fig:locomo_avg_token}
  \vspace{-12pt}
\end{figure}

\begin{table*}[]
\centering
\caption{Overall comparison across six tasks in LongMemEval-S benchmark: 
\textbf{Single-session-preference}, \textbf{Single-session-assistant}, \textbf{Temporal-reasoning}, \textbf{Multi-session}, \textbf{Knowledge-update}, and \textbf{Single-session-user}.
Reported F1 and Accuracy are multiplied by 100.}
\label{tab:overall_longmemeval-S_accuracy}
\resizebox{\textwidth}{!}{%
\begin{tabular}{l|ccc|ccc|ccc|ccc|ccc|ccc}
\toprule[1.5pt]
\multirow{2}{*}{Memory Module} & \multicolumn{3}{c|}{Single-session-preference} & \multicolumn{3}{c|}{Single-session-assistant} & \multicolumn{3}{c|}{Temporal-reasoning} & \multicolumn{3}{c|}{Multi-session} & \multicolumn{3}{c|}{Knowledge-update} & \multicolumn{3}{c}{Single-session-user} \\ \cline{2-19} 
 & F1 & Accuracy & LLM-as-a-Judge & F1 & Accuracy & LLM-as-a-Judge & F1 & Accuracy & LLM-as-a-Judge & F1 & Accuracy & LLM-as-a-Judge & F1 & Accuracy & LLM-as-a-Judge & F1 & Accuracy & LLM-as-a-Judge \\ \midrule[1pt]
ReadAgent & 3.54 & 4.17 & 1.25 & 4.48 & 15.18 & 1.46 & 3.76 & 4.32 & 0.86 & 1.65 & 4.89 & 1.03 & 2.96 & 8.01 & 1.05 & 4.87 & 17.14 & 1.55 \\
MemoryBank & 4.25 & 5.00 & 1.88 & 5.36 & 18.21 & 2.20 & 4.50 & 5.19 & 1.29 & 1.98 & 5.86 & 1.55 & 3.55 & 9.62 & 1.58 & 5.62 & 20.57 & 2.33 \\
MemGPT & 4.95 & 5.83 & 1.72 & 6.25 & 21.25 & 2.01 & 5.25 & 6.05 & 1.18 & 2.31 & 6.84 & 1.41 & 4.15 & 11.22 & 1.43 & 6.38 & 24.00 & 2.14 \\
A-mem & 7.78 & 9.17 & 2.50 & 9.86 & 33.39 & 2.93 & 8.28 & 9.51 & 1.72 & 3.64 & 10.75 & 2.06 & 6.54 & 17.63 & 2.10 & 10.03 & 37.71 & 3.10 \\
MemoryOS & 9.21 & 10.83 & 2.66 & 11.67 & 39.46 & 3.11 & 9.78 & 11.24 & 1.83 & 4.30 & 12.70 & 2.19 & 7.73 & 20.83 & 2.23 & 11.86 & 44.57 & 3.23 \\
MemOS & 10.61 & 12.50 & 2.81 & 13.39 & 45.53 & 3.29 & 11.23 & 12.97 & 1.94 & 4.94 & 14.66 & 2.31 & 8.88 & 24.04 & 2.36 & 13.63 & 51.43 & 3.29 \\ \midrule[1pt]
\sysname & 14.14 & 16.67 & \textbf{3.13} & 17.92 & 60.71 & \textbf{3.66} & 15.03 & 17.29 & \textbf{2.15} & 6.61 & 19.54 & \textbf{2.57} & 11.83 & 32.05 & \textbf{2.63} & 19.48 & 68.57 & \textbf{3.88} \\ \bottomrule[1pt]
\end{tabular}%
}
\end{table*}

\textbf{LoCoMo Analysis.}  
On the LoCoMo tasks, \sysname delivers strong retrieval accuracy that rivals or exceeds prior systems (Table~\ref{tab:overall_locomo_accuracy}).  
For example, on \textit{Temporal Reasoning}, \sysname achieves F1~$\approx 38.3$, substantially higher than the 26.5 F1 reported for MemGPT.  
Similarly, on \textit{Open-Domain}, \sysname attains 48.4 F1 compared to 41.5 for MemoryOS.  
\sysname’s LLM-as-a-Judge scores are also the highest in all categories ($\approx 3.0$–$3.3$ out of 5), reflecting answer quality that is equal or better than the baselines.  
These results demonstrate that the semantic-approximation mechanism in \sysname (binary token signatures) incurs only minor accuracy loss, while still retrieving relevant context effectively.  
In contrast, lightweight baselines like MemoryBank, which sacrifice search overhead for speed, achieve very low accuracy (F1~$<10$ across tasks).  
In summary, \sysname matches or outperforms SOTA memory systems on LoCoMo while using a compact, compressed-index representation.

The efficiency advantages of \sysname are dramatic.  
Figure~\ref{fig:locomo_search_ratio} plots the average end-to-end query latency for each system. 
\sysname responds in roughly $1.08$ seconds on average—an order of magnitude faster than dense-vector approaches (MemGPT~$\approx 33.6$s) and substantially quicker than knowledge graph- or RAG-based memories.  
The breakdown in Figure~\ref{fig:locomo_search_ratio} shows that \sysname spends only a small fraction of that time in the search phase,  
whereas baselines incur a dominant search cost (often $>80\%$ of total latency).  
Figure~\ref{fig:locomo_avg_token} displays average token consumption:  
\sysname reads only $\approx 1.3$K tokens on average, far fewer than MemGPT ($\approx 16.9$K) or MemoryOS ($\approx 8.1$K).  
This low token overhead arises from \sysname’s compressed memory structure: rather than loading large text embeddings, it scans concise bitwise signatures and reconstructs exact token IDs on demand.  
These efficiency gains show that \sysname achieves high retrieval accuracy with minimal latency and cost.
The observed performance aligns with our design motivation:  prior high-accuracy memory systems required heavy token usage and slow searches, whereas \sysname breaks that trade-off through its space-efficient, bitwise index.

\begin{figure}[!t]
\centering
  \includegraphics[width=0.95\columnwidth]{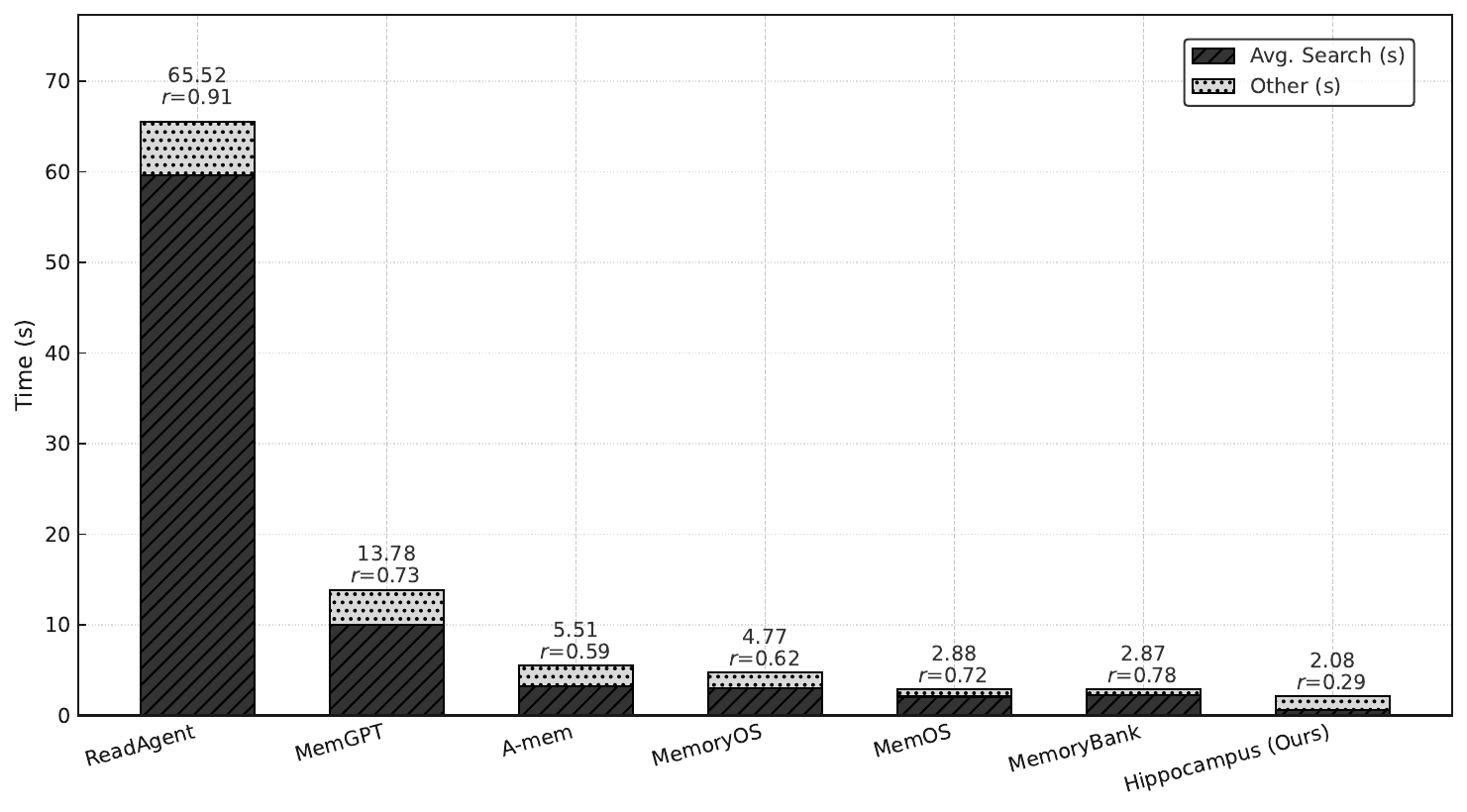}
  \caption{Average query retrieval latency (seconds) for various memory systems on LongMemEval-s.}
  \label{fig:longmemeval_s_latency}
  \vspace{-12pt}
\end{figure}
\begin{figure}[!t]
\centering
  \includegraphics[width=0.95\columnwidth]{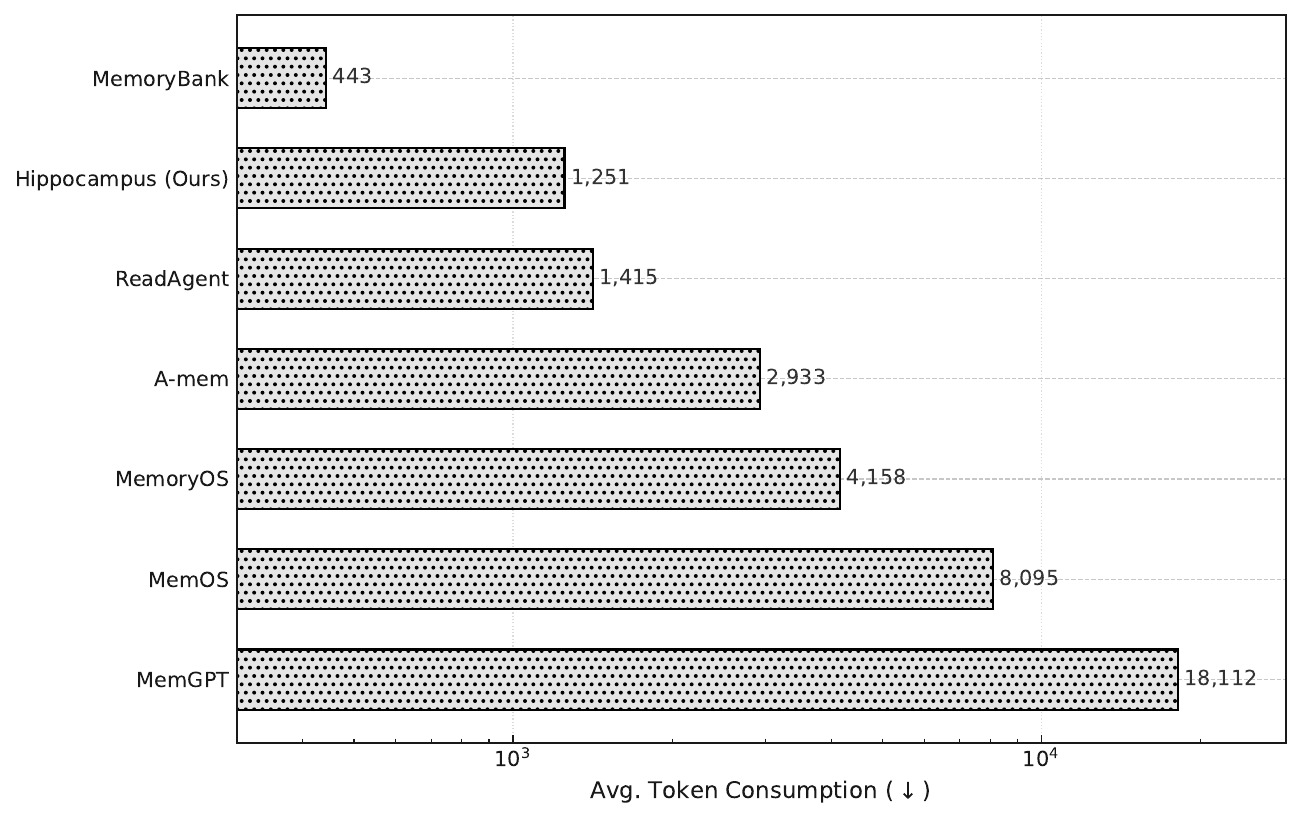}
  \caption{Average number of tokens consumption per query by each memory system on LongMemEval-s.}
  \label{fig:longmemeval_s_avg_token}
  \vspace{-16pt}
\end{figure}

\textbf{LongMemEval Analysis.}  
On LongMemEval-S, \sysname consistently achieves the best accuracy–efficiency operating point among all baselines.  
As summarized in Table~\ref{tab:overall_longmemeval-S_accuracy}, \sysname improves accuracy across all six tasks while dramatically reducing end-to-end retrieval time and minimizing token footprint.  
In Figure~\ref{fig:longmemeval_s_latency}, our end-to-end latency lies near the floor of the plot, reflecting how bit-sliced Hamming-ball filtering on the Signature DWM eliminates the dominant search cost that burdens dense-vector and graph-traversal designs.  
Figure~\ref{fig:longmemeval_s_avg_token} shows a much smaller per-query token budget, as \sysname scans compact binary signatures and reconstructs token IDs on demand,  
rather than streaming long textual passages or large embedding blocks.  
Together, these effects validate our design thesis from Section~\ref{sec: design}:  
approximate semantic access in the compressed domain (signatures) combined with exact reconstruction (Content DWM) breaks the classic trade-off—maintaining task accuracy while achieving order-of-magnitude gains in responsiveness and prompt-token economy.  
Please refer to Appendix~\ref{app: longmemeval-m} for complementary results on LongMemEval-M.

\section{Related Work}\label{sec: related_work}
The landscape of memory systems for agentic AI is rapidly evolving, with recent work focusing on high-level architectural abstractions to manage long-term experiences. These approaches can be broadly categorized into two dominant philosophies. The first draws inspiration from operating systems, treating memory as a manageable system resource. This includes MemGPT~\cite{packer2023memgpt}, which introduces virtual context management analogous to OS-level memory paging; MemoryOS~\cite{kang2025memory}, which implements a hierarchical storage architecture with short, mid, and long-term tiers; and MemOS~\cite{li2025memos}, which proposes a standardized MemCube abstraction to unify parametric, activation, and plaintext memory. The second category is inspired by human cognitive science, such as ReadAgent~\cite{lee2024human}, which compresses memories into gist memories~\cite{abadie2013gist} akin to human summarization; MemoryBank~\cite{zhong2024memorybank}, which employs an Ebbinghaus-inspired~\cite{tulving1985ebbinghaus} forgetting curve for dynamic memory updates; and A-mem~\cite{xu2025mem}, which organizes knowledge into an evolving, interconnected network based on the Zettelkasten method~\cite{malashenko2023digital}. Despite their architectural diversity, these systems converge on a common technological substrate where retrieval is predominantly powered by dense vector similarity search within a Retrieval-Augmented Generation (RAG) framework or by traversing explicit knowledge graph structures. For instance, A-mem leverages a vector store like ChromaDB
, and MemoryBank uses FAISS~\cite{douze2025faiss} for efficient retrieval. A more detailed related work is presented in Appendix~\ref{app: detailed_related_work}.

\vspace{-0.1in}
\section{Conclusion}\label{sec: conlusion}
This work presents \sysname, a contextual memory module that design with binary signatures and a Dynamic Wavelet Matrix co-index for compressed-domain search and lossless content reconstruction. The design scales linearly with history length, supports streaming writes, and executes semantic access via low-level bitwise primitives. Across LoCoMo and LongMemEval, \sysname preserves or improves task accuracy while substantially cutting both query latency and prompt-token cost, validating the effectiveness of approximate-then-exact retrieval in a succinct data structure.




\nocite{langley00}

\bibliography{example_paper}
\bibliographystyle{mlsys2025}

\clearpage
\appendix
\section{LLM-as-a-Judge Metric}\label{app: prompt}
When evaluating a generated answer, we feed both the reference and candidate into the judge prompt (see Listing~\ref{lst:judge-prompt}), and have GPT-5 act as the impartial judge to assign a score in the range of $[1,2,3,4,5]$. The LLM judge supplements F1, BLEU-1, and accuracy, which may overestimate correctness in the edge cases.

As shown in Listing~\ref{lst:judge-prompt}, we provide our prompt for LLM-as-a-Judge, which serves as the evaluation prompt for assessing the quality of generated answers. The prompt instructs an impartial evaluator to rate a candidate answer against a reference answer on a $[1,2,3,4,5]$ scale, focusing on \textbf{Correctness}, \textbf{Completeness}, and \textbf{Clarity/Coherence}. Specifically, a score of 5 indicates a perfectly correct, complete, and clear response; 4 reflects minor inaccuracies or slight omissions; 3 denotes partial correctness with missing major points; 2 corresponds to largely incorrect or irrelevant content; and 1 represents a completely wrong answer. This standardized prompt ensures consistent, interpretable, and reproducible evaluation across different experimental settings.
\begin{lstlisting}[language=Python, caption={Judge Prompt Template}, label={lst:judge-prompt}]
JUDGE_PROMPT = "
You are an impartial evaluator. 
Your task is to rate the quality of a
candidate answer compared to a reference answer.
[Question/Query]: {question};
[Reference Answer]: {reference};
[Candidate Answer]: {candidate}.
Please assign a score from 1 to 5 based on how well the {candidate} matches
the {reference} in terms of correctness, completeness (coverage of key points),
and clarity and coherence.

Scoring Guidelines:
5: perfectly correct, complete, and clear;
4: mostly correct, with minor issues or slight omissions;
3: partially correct, with noticeable errors or missing major points;
2: largely incorrect, irrelevant, or nonsensical;
1: totally wrong.

Only output the final score as an integer between 1 and 5.
"
\end{lstlisting}
As shown in Listing~\ref{lst:context-prompt}, we present the prompt for answering the question, which is used to instruct the model to generate answers strictly based on the provided context. The prompt explicitly constrains the model to avoid relying on external knowledge or prior training data, ensuring that the generated responses are fully grounded in the given information. By including placeholders for the context and question, this design enforces factual consistency and prevents hallucination, making it suitable for controlled evaluations of context-dependent reasoning and information retrieval tasks.
\begin{lstlisting}[language=Python, caption={Answer-from-Context Prompt}, label={lst:context-prompt}]
ANSWER_PROMPT = "
Based ONLY on the following context, answer the user's question directly.
Context:{context}
Question: {question}
"
\end{lstlisting}

\section{Detailed Dataset Description}\label{app: experimental_setting}
We use LoCoMo~\cite{maharana2024evaluating} and LongMemEval (-S and -M)~\cite{wu2024longmemeval} for the experiments. Below is the detailed description of these two benchmarks.
\begin{itemize}[topsep=0pt,itemsep=-1ex,partopsep=1ex,parsep=1ex, leftmargin=*]
\item LoCoMo is introduced to evaluate extremely long-term conversational memory in LLM agents. It is constructed via a machine-human hybrid pipeline: two LLM-powered agents carry multi-session dialogues grounded on persona profiles and temporal event graphs, generating coherent and causally linked conversations which humans then refine for consistency. Each conversation spans up to approximate 32 sessions and contains on the order of 600 turns and $\sim$16K tokens on average. The benchmark supports multiple tasks, including question answering~\cite{zaib2022conversational}, event summarization~\cite{kurisinkel2023llm}, and multimodal dialogue generation~\cite{liu2023matcr,huang2024dialoggen}, allowing evaluation along dimensions such as single-hop, multi-hop, temporal, and open-domain memory reasoning.
\item LongMemEval is a more recent benchmark tailored for chat assistants, designed to probe long-term memory in interactive, multi-session settings. It comprises 500 curated questions, each embedded within a dynamically constructed chat history spanning multiple sessions. The benchmark assesses five core memory abilities: information extraction~\cite{singh2018natural,peng2024metaie}, multi-session reasoning, temporal reasoning, knowledge updates, and abstention~\cite{xin2021art}. During evaluation, models must parse incremental interactions, maintain memory over sessions, and deliver answers after the final session, thereby simulating realistic real-world continual-memory demands.
\end{itemize}

\section{Overall Comparison on LongMemEval-M}\label{app: longmemeval-m}
\begin{table*}[]
\centering
\caption{Overall comparison of different memory modules across six tasks in LongMemEval-M benchmark, under the same setting as Table~\ref{tab:overall_longmemeval-S_accuracy}.}
\label{tab:overall_longmemeval-M_accuracy}
\resizebox{\textwidth}{!}{%
\begin{tabular}{l|ccc|ccc|ccc|ccc|ccc|ccc}
\toprule[1.5pt]
\multirow{2}{*}{Memory Module} & \multicolumn{3}{c|}{Single-session-preference} & \multicolumn{3}{c|}{Single-session-assistant} & \multicolumn{3}{c|}{Temporal-reasoning} & \multicolumn{3}{c|}{Multi-session} & \multicolumn{3}{c|}{Knowledge-update} & \multicolumn{3}{c}{Single-session-user} \\ \cline{2-19} 
 & F1 & Accuracy & LLM-as-a-Judge & F1 & Accuracy & LLM-as-a-Judge & F1 & Accuracy & LLM-as-a-Judge & F1 & Accuracy & LLM-as-a-Judge & F1 & Accuracy & LLM-as-a-Judge & F1 & Accuracy & LLM-as-a-Judge \\ \midrule[1pt]
ReadAgent & 3.45 & 0.83 & 1.15 & 2.72 & 5.36 & 1.45 & 3.17 & 1.69 & 1.04 & 1.20 & 1.32 & 1.08 & 1.56 & 2.57 & 1.19 & 2.47 & 6.79 & 1.06 \\
MemoryBank & 4.14 & 1.00 & 1.72 & 3.27 & 6.43 & 1.43 & 3.81 & 2.03 & 1.11 & 1.45 & 1.58 & 1.17 & 1.88 & 3.08 & 1.18 & 2.98 & 8.14 & 1.54 \\
MemGPT & 4.83 & 1.17 & 1.58 & 3.81 & 7.50 & 1.31 & 4.45 & 2.37 & 1.02 & 1.70 & 1.84 & 1.07 & 2.19 & 3.59 & 1.08 & 3.49 & 9.50 & 1.39 \\
A-mem & 7.59 & 1.83 & 2.30 & 5.99 & 11.79 & 1.91 & 7.00 & 3.72 & 1.49 & 2.68 & 2.89 & 1.57 & 3.46 & 5.64 & 1.58 & 5.49 & 14.93 & 1.94 \\
MemoryOS & 8.98 & 2.16 & 2.44 & 7.10 & 13.93 & 2.02 & 8.29 & 4.40 & 1.57 & 3.18 & 3.42 & 1.64 & 4.12 & 6.67 & 1.65 & 6.53 & 17.64 & 2.01 \\
MemOS & 10.36 & 2.50 & 2.58 & 8.21 & 16.07 & 2.14 & 9.59 & 5.08 & 1.65 & 3.68 & 3.95 & 1.72 & 4.77 & 7.70 & 1.73 & 7.56 & 20.36 & 2.16 \\ \midrule[1pt]
\sysname & 13.79 & 3.33 & \textbf{2.87} & 10.88 & 21.43 & \textbf{2.38} & 12.69 & 6.77 & \textbf{1.86} & 4.81 & 5.26 & \textbf{1.94} & 6.23 & 10.26 & \textbf{1.98} & 8.67 & 27.14 & \textbf{2.40} \\ \bottomrule[1pt]
\end{tabular}%
}
\end{table*}
\begin{figure*}[!t]
\centering
\includegraphics[width=0.95\linewidth]{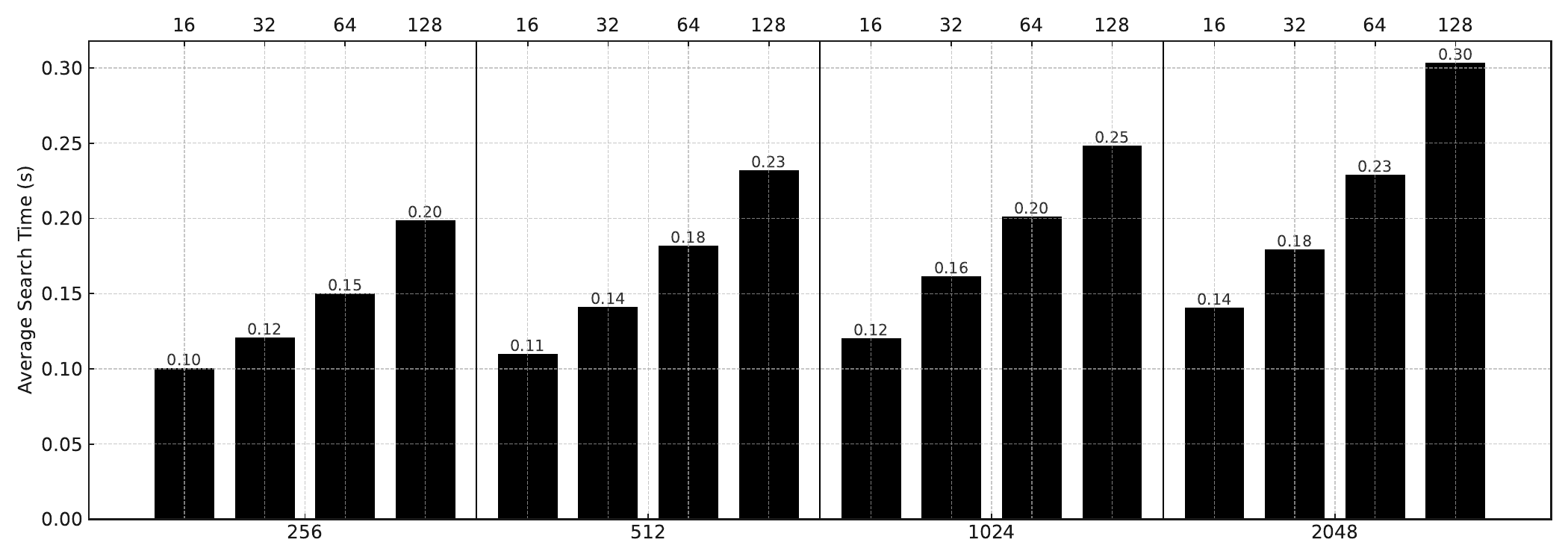}
  \caption{Ablation on the LoCoMo:
Average search time versus Random Indexing dimension $D$ and signature size $d$. 
Bars within each group (bottom axis) correspond to different $d$ values (top axis). }
  \label{fig:ablation_search}
  \vspace{-12pt}
\end{figure*}
\begin{table}[t]
\centering
\caption{LoCoMo ablation: F1/BLEU-1/LLM-as-a-Judge vs. Random Indexing $D$ and signature size $d$.}
\label{tab:ablation_locomo_accuracy}
\begin{tabular}{ll|ccc}
\toprule[1.5pt]
$D$ & $d$ & F1 (\%) & BLEU-1 (\%) & LLM-Judge \\
\midrule[1pt]
256 & 16 & 37.28 & 33.69 & 3.18 \\
256 & 32 & 38.27 & 34.53 & 3.17 \\
256 & 64 & 37.29 & 33.62 & 3.20 \\
256 & 128 & 37.54 & 34.20 & 3.20 \\ \midrule[1pt]
512 & 16 & 38.14 & 34.42 & 3.17 \\
512 & 32 & 38.30 & 33.78 & 3.18 \\
512 & 64 & 37.75 & 33.45 & 3.17 \\
512 & 128 & 38.16 & 34.42 & 3.21 \\ \midrule[1pt]
1024 & 16 & 38.72 & 33.83 & 3.20 \\
1024 & 32 & 38.18 & 34.49 & 3.20 \\
1024 & 64 & 37.31 & 33.63 & 3.19 \\
1024 & 128 & 37.59 & 33.46 & 3.19 \\ \midrule[1pt]
2048 & 16 & 37.79 & 33.82 & 3.18 \\
2048 & 32 & 38.66 & 34.20 & 3.21 \\
2048 & 64 & 38.35 & 33.54 & 3.19 \\
2048 & 128 & 38.21 & 34.08 & 3.21 \\
\bottomrule[1.5pt]
\end{tabular}
\end{table}

Accuracy-only results on LongMemEval-M, as shown in Table~\ref{tab:overall_longmemeval-M_accuracy}, mirror the trends observed on the LongMemEval-S (Table~\ref{tab:overall_longmemeval-S_accuracy}): \sysname attains the highest or near-highest accuracy across all categories, particularly on multi-session and knowledge-update where signature-level association helps surface temporally and semantically related memories that are dispersed across sessions. We omit efficiency plots for brevity, the latency and token-consumption advantages follow the same pattern as on LongMemEval-S, since the retrieval substrate (signature filtering and content reconstruction) is identical; hence the relative gaps against dense-vector and KG baselines persist at similar magnitudes.

\section{ablation Study of \sysname}\label{app: breakdown}
We evaluated expected trends on the LoCoMo long-memory benchmark by varying the random-index dimension $D\in \{256,512,1024,2048\}$ and binary signature length $d \in \{16,32,64,128\}$. Figure~\ref{fig:ablation_search} reports the expected average search time (in seconds), and Table~\ref{tab:ablation_locomo_accuracy} shows quality metrics (F1, BLEU-1, and LLM-as-a-Judge) under these settings.\\
\textbf{Avg. Search Time.} Retrieval involves computing Hamming distances between the query’s binary signature and all stored signatures. Thus search time grows roughly linearly with the signature length $d$ (and weakly with $D$). In our table, doubling $d$ roughly doubles the time. Each extra bit adds a fixed cost, so larger $d$ or $D$ slows lookup.\\
\textbf{Accuracy.} Increasing $D$ or $d$ raises the representational capacity of the memory, reducing collisions and improving recall. A higher random indexing dimension $D$ yields more nearly-orthogonal random codes, while a longer signature $d$ captures more bits of information. Consequently, all quality metrics (F1, BLEU-1, and the LLM-as-a-Judge score) improve as $D$ and $d$ grow. This matches the known trends: expanding memory capacity or embedding dimensions consistently boosts retrieval performance~\cite{li2025memos}.\\
\textbf{Trade-off Consideration.} There is a clear trade-off. Larger $D$ and $d$ yield diminishing marginal gains in accuracy (the improvements taper off as the system saturates its capacity), but each added dimension/bit linearly increases search effort. In practice, one chooses $D$, $d$ to balance these effects: enough capacity to achieve good recall accuracy (and thus higher LLM-judge scores), but not so large that retrieval becomes too slow.

\section{Memory Construction}\label{app: construction}
A practical memory system must not only support fast retrieval at inference time, but also allow efficient memory construction (i.e., ingesting raw content into a persistent memory substrate). This cost directly impacts usability in real deployments, where memories are frequently refreshed, re-indexed, or rebuilt under updated policies.

We measure the end-to-end wall-clock time to construct memory on LoCoMo. The measurement starts from reading the raw LoCoMo records and ends when the memory store is fully materialized and ready for querying (including all preprocessing, indexing, and persistence steps required by each method). We also report the total LLM token consumption (base model is gpt-4o-mini) incurred during construction, which captures the overhead of LLM-based summarization, or rewriting pipelines. All methods are evaluated under the same hardware/software environment.

Table~\ref{tab:locomo-construction} shows that Hippocampus constructs memory in only 6.70 minutes while consuming zero LLM tokens. In contrast, prior systems rely on LLM-intensive preprocessing (e.g., summarization or memory rewriting) and thus incur substantial token usage and significantly higher wall-clock time. Compared with the fastest baseline in this table (A-mem), Hippocampus is 5.3$\times$ faster, while eliminating token costs entirely. This advantage stems from our embedding-free construction pipeline based on token-id streams and binary signatures, avoiding LLM calls during ingestion.

\section{Time and Space Complexity}\label{app: dwm_time_space}
\begin{table}[]
\centering
\caption{End-to-end memory construction cost on LoCoMo. We report wall-clock time and total token consumption.}
\label{tab:locomo-construction}
\resizebox{\columnwidth}{!}{%
\begin{tabular}{l|c|c}
\toprule[1.5pt]
Method             & Time (minute) & Token Consumption \\ \midrule[1pt]
MemoryOS           & 4458.96       & 41540             \\
Nemori             & 477.66        & 27637             \\
A-mem              & 35.69         & 19926             \\
MemGPT             & 59.49         & 50674             \\
MemOS              & 70.00         & 21055             \\ \midrule[1pt]
Ours (Hippocampus) & 6.70          & 0                 \\ \bottomrule[1.5pt]
\end{tabular}%
}
\end{table}
We compare the asymptotic efficiency of \sysname’s DWM-based storage/retrieval against dense-vector search (e.g. FAISS) and knowledge-graph methods. 

Let $n$ be the number of tokens stored (i.e., total memory size) and let $\sigma$ be the dictionary size. Let each binary signature have length $d$ bits:\\
\textbf{DWM Construction (memory insertion).} Each token-id insertion into the DWM (both content and signature matrices) requires a top-down traversal of $l=\lceil\log_2\sigma\rceil$ bit-level wavelet levels. Each level uses rank/select on a dynamic bitvector. In a static wavelet matrix, rank is $\mathcal{O}(1)$ with succinct overhead, in a dynamic setting rank/select takes $\mathcal{O}(\log n)$ time per operation. Hence a single append costs $\mathcal{O}(\ell + \log n)=\mathcal{O}(\log\sigma +\log n)$. Amortized over $n$ inserts, total build time is $\mathcal{O}(n\log n)$ (dominated by dynamic updates). Space is $nl+o(nl)$ bits (i.e. $\mathcal{O}(n\log\sigma)$ bits) for the raw bit-matrix, plus overhead for rank/select. In summary, DWM insertion is nearly linear: $\mathcal{O}(n\log n)$ time and $\mathcal{O}(n\log\sigma)$ space.\\
\textbf{DWM Query (retrieval).} Exact pattern matching in DWM (finding a specific signature) takes $\mathcal{O}(l)$ time via rank/select (Algorithm~\ref{alg:dwm-rank}). However, \sysname performs an approximate Hamming ball search. This is done by scanning each stored signature: for each candidate signature bit-string $s_i$, we compute $\mathrm{Ham}(s_q,s_i)$ by bitwise XOR and popcount. Using machine words of size $w$ (e.g. $w=64$), each signature takes $\mathcal{O}(d/w)$ bit-operations. Thus the search cost is $\mathcal{O}(n\cdot d/w)$, i.e. linear in $n$ (and linear in $d$ bitwise operations). In practice $d$ is modest (e.g. a few hundred) so this is efficient, but asymptotically still $\mathcal{O}(n)$. In contrast, dense-vector search is also linear in $n$ in the worst case (even using indexing), but often with a larger constant due to dimensional inner products.\\
\textbf{Dense-vector ANN (e.g. FAISS).} A brute-force k-NN search in $D$-dimensional space costs $\mathcal{O}(nD)$ per query. Modern systems use specialized indices (product quantization, HNSW graphs) to achieve sublinear query time on average. For example, HNSW scales roughly as $\mathcal{O}(\log n)$ queries for well-behaved data, but has worst-case cost $\mathcal{O}(n)$. The Faiss library~\cite{douze2025faiss} implements a variety of indexes, but fundamental limits remain: in the worst case, retrieval requires examining many candidates. Space overhead for Faiss indexes is typically $\mathcal{O}(nD)$ to store the vectors plus index overhead.\\
\textbf{Knowledge Graph Traversal.} If memories are stored as a knowledge graph of entities and relations, a query may involve multi-hop neighbor exploration. In general, a breadth-first search up to $h$ hops from a starting node touches $\mathcal{O}(\Delta^h)$ nodes, where $\Delta$ is the average branching factor. In practice, $h$ is kept small (e.g. 2–3), but even then exploring the graph can be expensive. In worst-case terms, a multi-hop query is $\mathcal{O}(|V|+|E|)$, where $V$ is node count and $E$ edges. Storing a full graph also uses $\mathcal{O}(|V|+|E|)$ space. Notably, dynamic DWM updates and searches avoid such combinatorial growth.\\
\textbf{Comparison.} Asymptotically, \sysname’s DWM has linear space and near-linear-time updates ($\mathcal{O}(n\log n)$ build, dominated by dynamic bit-vector updates. Retrieval is also $\mathcal{O}(n)$ per query (with a small bit-level factor). Dense-vector methods typically require $O(nD)$ worst-case and may need $\mathcal{O}(n\log n)$ pre-processing. Graph methods can suffer exponential blow-up in hops or at least $\mathcal{O}(n)$ per query. Thus, in theory the DWM+Hamming search is comparable or better than naive baselines and avoids the multi-hop expansion cost of graphs. Practically, the use of bitwise operations and in-memory bit-slices makes \sysname much faster per comparison than dense multiplications, as confirmed by our experiments.

\section{Accuracy Gap}\label{app: acc_gap}
We formalize how \sysname’s random indexing step followed by an $r$-bit Hamming ball search approximates dense-vector similarity. Let each token’s context embedding be a vector $\bm{v}\in\mathbb{R}^D$, and consider two such vectors $\bm{v},\bm{w}$. \sysname generates a $d$-bit signature by random projection and thresholding: each bit is $\text{sign}(\langle b_k, v\rangle)$ for some random hyperplane $b_k$ (or an analogous sparse random base-vector scheme). By known results for random hyperplane hashing, the probability that a single bit differs satisfies:
$$
P(\text{bit}_k(\bm{v})\neq \text{bit}_k(\bm{w})) = \frac{\theta}{\pi}
$$
where $\theta=\arccos \big(\frac{\bm{v}\cdot \bm{w}}{\vert \bm{v}\vert \vert \bm{w} \vert}\big)$. Thus the expected Hamming distance between the $d$-bit signatures is
$$
\mathbb{E}[\text{Ham}(\bm{v},\bm{w})]=d\cdot \frac{\theta}{\pi}
$$
Equivalently, similarity $\frac{\bm{v}\cdot \bm{w}}{\vert \bm{v} \vert \vert \bm{w} \vert} = \cos\theta$ can be recovered up to small error from the normalized Hamming similarity.

With $d$ independent bits, the law of large numbers gives concentration: for any $\varepsilon>0$, by Hoeffding’s bound:
$$
P(\vert \frac{1}{d}\text{Ham}(\bm{v},\bm{w})-\frac{\theta}{\pi}\vert \geq \epsilon) \leq 2e^{-2d\epsilon^2}
$$
Hence with high probability, $\frac{1}{d}\text{Ham}(\bm{v},\bm{w})$ is within $\mathcal{O}(1/\sqrt{d})$ of $\theta/\pi$. In practice, choosing $d=\mathcal{O}(\epsilon^{-2}\log N)$ ensures that for any fixed query among $N$ candidates, the Hamming distance will approximate the original cosine similarity within additive error $\epsilon$.

Concretely, if we set a Hamming threshold $r$ corresponding to a desired angle $\theta_0$ (thus target similarity $\cos\theta_0$), then any $\bm{w}$ with $\frac{\bm{v}\cdot \bm{w}}{|\bm{v}||\bm{w}|} \ge \cos\theta_0$ will satisfy $\text{Ham}(\bm{v},{w})\le r$ except with probability at most $e^{-\mathcal{O}(d)}$. Conversely, vectors with similarity below $\cos\theta_0$ will exceed the threshold with high probability. This establishes that \sysname’s random indexing plus Hamming ball filter retrieves all sufficiently similar vectors (within angle $\theta_0$) with bounded false-negative probability, and rejects dissimilar vectors, mirroring an approximate nearest-neighbor search in cosine similarity space. The sampling complexity matches known bounds for binary embeddings.\\
\textbf{Theorem.} Under the process in Section~\ref{sec: RI_HB}, for any two vectors $\bm{v},\bm{w}$, the Hamming distance of their $d$-bit signatures concentrates around its mean $d\theta/\pi$. By choosing $d=\mathcal{O}(\delta^{-2}\log(1/\eta))$, one ensures $\text{Ham}(\bm{v},\bm{w})/d$ approximates $\theta/\pi$ within $\pm\delta$ with probability $1-\eta$. In particular, setting the Hamming radius $r=\frac{d\theta_0}{\pi}$, the Hamming-ball ${s: \text{Ham}(\bm{s}_v,\bm{s})\le r}$ contains all items with cosine similarity at least $\cos(\theta_0)$ up to vanishing error.\\
\textbf{Proof.} \sysname compresses high-dimensional embeddings into fixed‐length binary signatures by sparse random indexing and binarization. In effect, each bit of a token signature can be viewed as the sign of a random hyperplane dot-product with the original vector. The Hamming distance between two signatures then equals the number of bits on which they differ. We will show that this Hamming distance concentrates around $(\theta/\pi)d$, where $\theta=\arccos\frac{\bm{v}\cdot \bm{w}}{|\bm{v}||\bm{w}|}$ is the angle between vectors $\bm{v},\bm{w}$. In particular, by choosing a suitable radius $r\approx(\theta_0/\pi)d$ we retrieve all vectors with angle $\le\theta_0$ (cosine similarity $\ge\cos\theta_0$) with high probability.\\
\textbf{Binary hash and collision probability.} For each bit index $i=1,\dots,d$, pick an independent random Gaussian vector $\bm{r}_i\sim N(0,I)$ in $\mathbb R^n$ and define the bit $b_i(v)=\operatorname{sign}(\bm{r}_i\cdot \bm{v})\in{0,1}$. (Equivalently, \sysname selects a sparse random base vector and later binarizes the top-$d$ components, which yields the same analysis.) Let $X_i$ be the indicator that $b_i(v)\neq b_i(w)$ (a bit mismatch). It is a known fact~\cite{charikar2002similarity} that for any two vectors $\bm{v},\bm{w}$ at angle $\theta$, the probability their signs agree on a random hyperplane is
$$
P(b_{i}(v)=b_{i}(w)) = 1 - \frac{\theta}{\pi} \Rightarrow P(X_{i} = 1) = \frac{\theta}{\pi}
$$
considering a random line in the 2D plane of $\bm{v},\bm{w}$ Thus $X_i\sim\mathrm{Bernoulli}(p)$ with $p=\theta/\pi$, and the Hamming distance $H(\bm{v},\bm{w})=\sum_{i=1}^d X_i$ is a binomial random variable with mean
$$
\mathbb{E}[H(\bm{v},\bm{w})] = dp=\frac{\theta}{\pi}d
$$
\textbf{Concentration (Hoeffding/Chernoff bound).} The $X_i$ are independent and bounded in $[0,1]$, so by Hoeffding’s inequality, we have for any $\epsilon>0$:
$$
P(\vert H(\bm{v},\bm{w})-dp\vert \geq \epsilon d) \leq 2e^{-2\epsilon^2 d}
$$
Equivalently:
$$
P(\vert \frac{H(\bm{v},\bm{w})}{d}-p\vert \geq \epsilon) \leq 2e^{-2\epsilon^2 d}
$$
Hence with high probability $H(v,w)/d$ lies in the interval $[p-\epsilon,;p+\epsilon]$, i.e.
$$
H(\bm{v},\bm{w}) = (\frac{\theta}{\pi})d \pm \epsilon d \quad \text{with probability} 1-2e^{-2\epsilon^2 d}
$$
\textbf{Containment in the Hamming ball (false negatives).} Fix a target angle $\theta_0$ (so we want $\cos(\bm{v},\bm{w})\ge\cos\theta_0$). Consider any vector $w$ with $\theta(\bm{v},\bm{w})\le\theta_0$. Then $p=\theta/\pi \le \theta_0/\pi$. Define a search radius: $r=(\frac{\theta_0}{\pi}+\epsilon)+d$, by the above tail bound:
$$
P(H(\bm{v},\bm{w}) > r) = P(\frac{H}{d}>\frac{\theta_0}{\pi} + \epsilon) \leq e^{-2\epsilon^2 d}
$$
since $\mathbb{E}[H/d]\le \theta_0/\pi$. Thus with probability at least $1-e^{-2\epsilon^2d}$ we have $H(v,w)\le r$. By choosing $d$ large enough (see below), this failure probability can be made $\le\delta/N$ (union-bounding over $N$ candidates). In summary, any vector within angle $\theta_0$ will lie inside the Hamming ball of radius $r\approx(\theta_0/\pi)d$ with high probability.\\
\textbf{False positives (outside angle).} Conversely, if $\theta(\bm{v},\bm{w}) > \theta_0$, then $p = \theta/\pi > \theta_0/\pi$. In particular, if $\theta\ge \theta_0+2\epsilon\pi$ then $p\ge\theta_0/\pi+2\epsilon$. In that case:
$$
P(H(\bm{v},\bm{w}) \leq (\frac{\theta_0}{\pi}+\epsilon)d) = P(\frac{H}{d}\leq p-\epsilon) \leq e^{-2\epsilon^2d}
$$
by the lower-tail Hoeffding bound. Hence vectors with angle substantially above $\theta_0$ will (with probability $1-\exp(-2\epsilon^2d)$) have Hamming distance exceeding $(\theta_0/\pi+\epsilon)d$ and will not be included in the ball of radius $r=(\theta_0/\pi+\epsilon)d$. This bounds the false‐positive rate.\\
\textbf{Parameter choice ($d$ vs. $\epsilon,\delta,N$).} To guarantee both error probabilities $e^{-2\epsilon^2d}$ are at most $\delta/(2N)$ (so that a union bound over $N$ vectors still yields failure probability $\le\delta$), it suffices to choose $d\geq \frac{1}{2\epsilon^2}\ln(\frac{2N}{\delta})$. In big‐O terms, $d=O\bigl((1/\epsilon^2)(\log N + \log(1/\delta))\bigr)$ is enough. For such $d$, we have with probability $1-\delta$ (over the randomness of the projections) that all vectors within angle $\theta_0$ lie in the Hamming ball of radius $r=(\theta_0/\pi+\epsilon)d$, and vectors with angle significantly larger than $\theta_0$ lie outside this ball.

The above calculation shows that the normalized Hamming distance $H(\bm{v},\bm{w})/d$ concentrates near $(\theta/\pi)$. Hence a Hamming‐ball query of radius $r\approx(\theta_0/\pi)d$ retrieves exactly those vectors with angle $\le\theta_0$ (cosine $\ge\cos\theta_0$), up to an error margin controlled by $\epsilon,\delta$. In other words, \sysname’s random indexing plus Hamming ball method yields an $(\epsilon,\delta)$-approximation of cosine-similarity search: with $d=O((1/\epsilon^2)\log(N/\delta))$ bits, one finds all high-cosine neighbors with bounded false-positive/negative rates

\section{Detailed Related Work}\label{app: detailed_related_work}
\subsection{Memory System for Agentic AI}
The design of memory modules for agentic AI has become a central research area, with a primary focus on developing high-level architectural frameworks that enable agents to store, organize, and recall past experiences effectively. A close examination of the state-of-the-art reveals that innovation has largely concentrated on the conceptual layer of memory management, how an agent should reason about its history, while relying on a common set of underlying retrieval technologies.

A prominent school of thought approaches agent memory by drawing analogies to the memory management principles of traditional operating systems (OS), emphasizing concepts like hierarchy, resource allocation, and control flow.\\
MemGPT~\cite{packer2023memgpt} pioneers the concept of virtual context management for LLMs. This technique provides the illusion of an infinite context window by creating a two-tiered memory hierarchy. The main context is analogous to physical RAM and consists of the tokens directly within the LLM's prompt, while the external context serves as disk storage for out-of-context information. The core mechanism of MemGPT is that the LLM itself orchestrates the movement of data between these tiers through self-directed function calls, effectively managing its own limited context as a constrained resource.\\
MemoryOS~\cite{kang2025memory} extends this OS metaphor with a more rigidly defined three-tier hierarchical storage architecture: Short-Term Memory (STM) for real-time conversations, Mid-Term Memory (MTM) for topic-based summaries, and Long-term Personal Memory (LPM) for persistent user and agent personas. It formalizes the data lifecycle with explicit update policies borrowed from OS design, such as a dialogue-chain-based FIFO (First-In, First-Out) principle for promoting information from STM to MTM and a heat-based replacement strategy for archiving less relevant information from MTM, mirroring OS page management techniques.\\
MemOS~\cite{li2025memos} presents the most abstract and comprehensive OS-level vision, proposing that memory should be treated as a first-class operational resource within the AI system. Its central innovation is the MemCube, a standardized data structure and abstraction layer designed to unify three fundamentally different memory types: parametric memory (knowledge encoded in model weights), activation memory (transient states like the KV-cache), and plaintext memory (external knowledge sources). By providing a unified framework for the full lifecycle of these memory units, including their creation, scheduling, and evolution, MemOS aims to imbue LLMs with system-level controllability, plasticity, and evolvability.

A second major approach draws inspiration from psychology and cognitive science, seeking to build memory systems that emulate the more nuanced and adaptive characteristics of human memory.\\
ReadAgent~\cite{li2023interactive} is modeled on how humans read and comprehend very long documents. Instead of attempting to process entire texts verbatim, it implements a system that creates short, compressed gist memories. This design is grounded in the fuzzy-trace theory of human memory, which posits that humans quickly forget precise details but retain the core substance or gist of information for much longer. ReadAgent uses the LLM's own reasoning capabilities to decide what content to group into a memory episode, how to compress it into a gist, and when to perform an interactive look-up of the original text for specific details, transforming retrieval into an active reasoning task.\\
MemoryBank~\cite{zhang2024interactive} explicitly incorporates a model of human forgetting to achieve more natural long-term interactions. Its memory update mechanism is directly inspired by the Ebbinghaus Forgetting Curve theory, a psychological principle describing the decay of memory over time. This allows the agent to selectively forget less significant or infrequently accessed memories while reinforcing more important ones, aiming for a more anthropomorphic and engaging user experience, particularly in long-term AI companion scenarios. Its storage is also hierarchical, distilling verbose dialogues into concise daily summaries, which are then aggregated into a global summary.\\
A-mem~\cite{xu2025mem} is architected around the principles of the Zettelkasten method, a sophisticated technique for knowledge management that emphasizes the creation of a network of interconnected atomic notes. When a new memory is formed, A-mem uses an LLM to generate a structured note containing attributes like keywords, tags, and a rich contextual description. The system then agentically analyzes historical memories to establish meaningful links, creating an evolving web of knowledge. This process also enables memory evolution, where the integration of new information can trigger updates to the attributes of existing memories, allowing the network to continuously refine its understanding over time.

A distinct architectural approach structures memory explicitly as a knowledge graph (KG), which excels at representing the relational and temporal dependencies between entities. While many systems rely on vector search for amorphous semantic similarity, KGs provide a structured representation that is particularly well-suited for tasks requiring multi-hop reasoning or a precise understanding of how the information evolves.\\
Zep and Graphiti~\cite{rasmussen2025zep} exemplify this approach. Zep is a memory layer service for agents that is powered by Graphiti, a temporally-aware knowledge graph engine. Unlike static RAG systems that retrieve from unchanging document collections, Graphiti dynamically ingests and synthesizes both unstructured conversational data and structured business data into a KG that explicitly maintains historical relationships and their periods of validity. This bi-temporal model, which tracks both event time and transaction time, enables agents to perform complex temporal reasoning queries (e.g., "What was the status of Project X last week?"), a capability that is fundamentally challenging for standard vector-based RAG systems.

\subsection{Succinct Data Structures}
The core data structure of \sysname: Dynamic Wavelet Matrix, is rooted in the field of succinct data structures, a specialized area of computer science focused on high-performance information retrieval in space-constrained environments.

Succinct data structures are data representations that occupy an amount of space that is very close to the information-theoretic minimum required to store the data, while still supporting efficient queries. For example, a binary tree with $n$ nodes requires at least $2n$ bits to be represented uniquely, and succinct representations achieve this bound while still allowing for navigation operations (e.g., finding a parent or child) in constant time. A crucial feature that distinguishes them from simple compression algorithms is that they are designed to be queried directly in their compressed form, without needing to be decompressed first. This combination of extreme space efficiency and fast query performance makes them ideal for managing massive datasets that must be held in memory. This field has matured from purely theoretical results to practical, highly-engineered libraries such as SDSL. 

The Wavelet Matrix is a powerful and flexible succinct data structure designed to represent long sequences of symbols, such as a stream of integers drawn from a fixed alphabet. It is an optimized and more practical implementation of the conceptual Wavelet Tree. Structurally, it reorganizes the bits of the symbols in the input sequence into a collection of bit-vectors, where each bit-vector corresponds to a specific bit-plane of the alphabet (e.g., the most significant bits of all symbols form the first bit-vector, the second-most significant bits form the second, and so on).\\
By augmenting these bit-vectors with small auxiliary structures that allow for constant-time binary \textit{rank} and \textit{select} operations, the Wavelet Matrix can efficiently support three fundamental queries on the original sequence in time logarithmic in the alphabet size ($\mathcal{O}(log~\sigma)$):\\
\textbf{access($i$)}: Returns the original symbol at position $i$;\\
\textbf{rank($c,i$)}: Counts the number of occurrences of symbol $c$ in the prefix of the sequence up to position $i$.\\
\textbf{select($c,j$)}: Finds the position of the $j$-th occurrence of symbol $c$ in the sequence.\\
These primitives are the computational building blocks used by \sysname. However, canonical wavelet matrices are static, they are built once over a fixed dataset and do not support efficient updates. 

A key technical contribution of our work is the development of a Dynamic Wavelet Matrix (DWM), an append-friendly adaptation specifically designed to handle the high-throughput, continuously growing memory stream of an agentic system. Furthermore, the application of this structure to co-index two heterogeneous data streams: compact semantic signatures for search and lossless token-IDs for reconstruction, is a novel use case that extends the traditional application of wavelet matrices in information retrieval.

\end{document}

%% file: 1_intro_v1.1.tex
\section{Introduction}
Agentic AI represents a transformative shift in how intelligent systems interact with the real-world. 
Unlike traditional software, which executes predefined logic, agentic systems autonomously perceive, plan, act, and adapt over time. 
Powered by large language models (LLMs)~\cite{vaswani2017attention,touvron2023llama,team2023gemini,liu2024deepseek}, these agents can decompose the complex tasks, invoke external tools, reflect on their own behavior, and revise strategies—all without explicit human supervision. 
Early prototypes such as AutoGPT~\cite{chen2023autoagents} and BabyAGI~\cite{nakajima2023babyagi} demonstrated that coupling the LLMs with goal-driven loops unlocks emergent capabilities far beyond static prompting.
As agentic AI transitions from research novelty to production infrastructure, it is poised to reshape productivity tools, DevOps workflows~\cite{ali2024optimizing}, and knowledge systems~\cite{zhu2024knowagent}. 

Memory is a core component of agentic AI systems. 
While perception and planning enable agents to respond to immediate stimuli, memory allows them to accumulate experience, maintain coherence across interactions, and reason over long temporal horizons. 
In practice, agentic systems operate within an \textit{observe–plan–act–learn} loop~\cite{srivastava2019sense,hayes1979cognitive}, where memory serves as the persistent substrate connecting past observations to future decisions. 
Without reliable recall, even sophisticated planners—such as those based on ReAct~\cite{yao2023react} or GoalAct~\cite{chen2025enhancing}—can lose track of prior actions, repeat failed strategies, or misinterpret context~\cite{li2023interactive,xu2025mem}. 
This limitation is exacerbated by the bounded context window of LLMs~\cite{su2024roformer,wu2024extending}, which restricts the amount of information that can be considered during inference. 
Empirical studies, including the “\textit{Lost-in-Middle}” effect~\cite{liu2023lost}, show that reasoning accuracy degrades sharply as prompts grow longer. 
As a result, context engineering~\cite{mei2025survey, context2025langchain} has emerged as a workaround, treating the context window as a scarce computational resource. 
However, this approach is brittle and labor-intensive. 
A dedicated memory management system offers a principled alternative: by externalizing long-term knowledge, it enables agents to retrieve only the most relevant fragments of prior information, e.g., dialogue, tool outputs, or retrieved facts, preserving context for immediate reasoning while maintaining continuity across tasks. 
Major agentic AI frameworks, e.g., LangChain and CrewAI~\cite{chase2022langchain, memory2025crewai}, already provide memory management systems to enhance agent capabilities, and real-world deployments report improved coherence and personalization when memory is enabled~\cite{li2023camel}.
\begin{figure}[!t]
\centering
  \includegraphics[width=0.9\columnwidth]{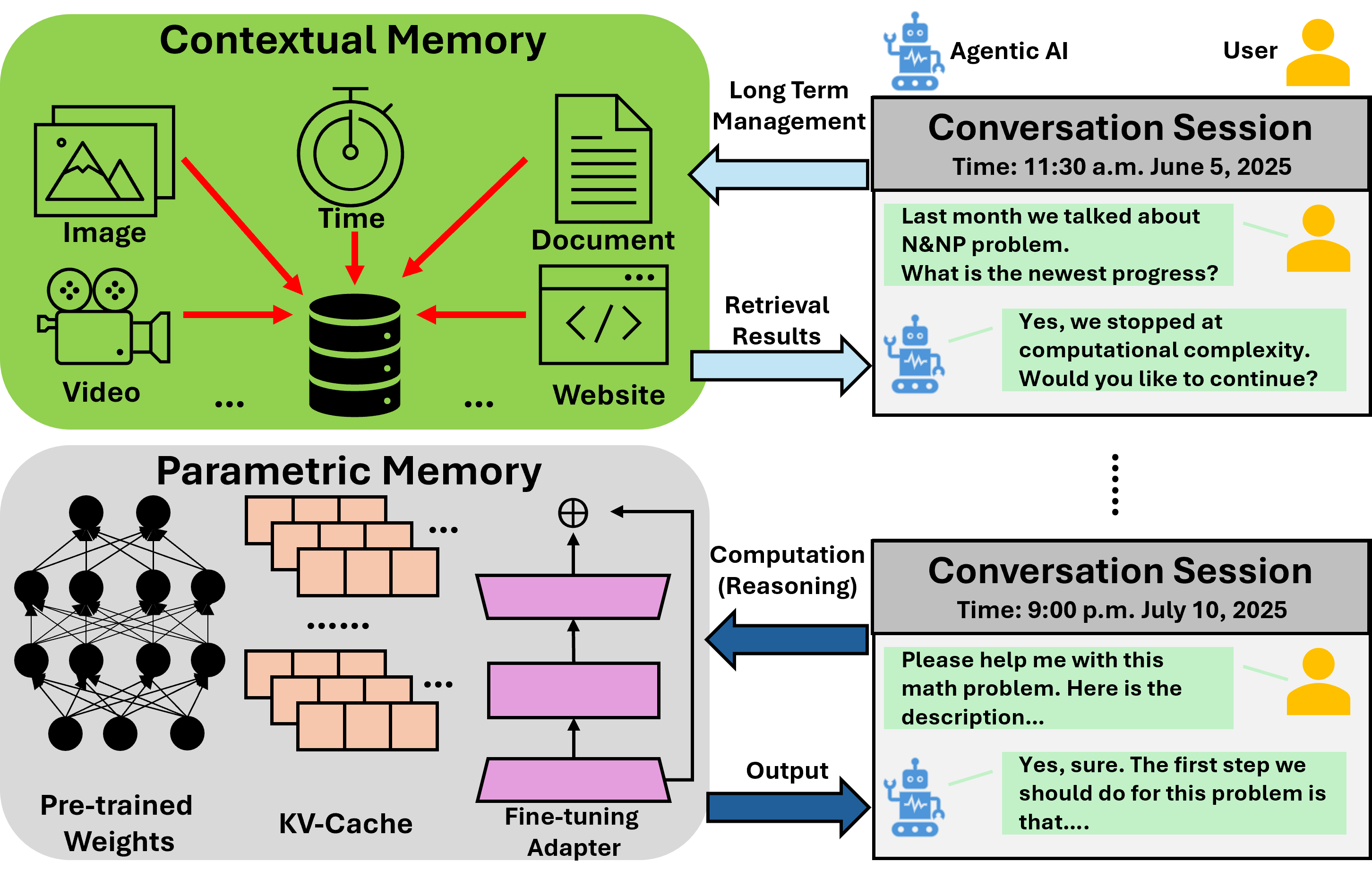}
  \caption{An illustration of memory taxonomy of Agentic AI.}
  \label{fig:para_context}
  \vspace{-12pt}
\end{figure}

Agentic memory can be broadly categorized into two types: \textit{parametric} and \textit{contextual}~\cite{du2025rethinking} (as shown in Figure~\ref{fig:para_context}). 
Parametric memory is embedded within the LLM itself—encoded in its weights, caches, or adapter layers~\cite{hu2022lora}. 
While powerful, it is opaque, expensive to update, and tightly coupled to model internals~\cite{wang2024knowledge}. 
In contrast, contextual memory is external and accessed via explicit retrieval, enabling agents to store and query long-term histories, tool outputs, and retrieved knowledge. 
This externalization offers three key advantages: 
(1) effectively unbounded capacity~\cite{de2021editing,jiang2024learning}; 
(2) fast, selective updates without retraining; and 
(3) schema-level interpretability and control. 
In this work, we focus on the contextual memory, which has emerged as a critical enabler for scalable, coherent, and responsive agentic AI systems.

Despite architectural diversity, existing contextual memory systems share a critical limitation: low efficiency in both memory insertion and retrieval. 
Whether based on RAG~\cite{zhang2024interactive,kagaya2024rap}, knowledge graph~\cite{kim2024leveraging,anokhin2024arigraph}, or hybrid designs, these systems incur substantial overhead when storing new memory entries and retrieving relevant content. 
Insertion often requires costly embedding generation and preprocessing, while retrieval relies on high-dimensional similarity search or multi-hop graph queries—both computationally expensive and latency-prone. 
This inefficiency is especially problematic in agentic AI, where agents operate in iterative loops and frequently update or consult memory across steps. 
Slow memory operations stall the observe–plan–act–learn cycle, reducing agent throughput and responsiveness. 
As agents scale to longer horizons and more complex tasks, the need for a memory substrate that supports fast, streaming writes and low-latency recall becomes paramount.
A detailed performance analysis is presented in section~\ref{sec: performance}.

To address above mentioned limitations, we argue for a fundamentally different memory substrate—one that abandons token-centric, embedding-heavy representations in favor of lightweight, compression-native structures. 
The goal is to support efficient memory insertion and retrieval without sacrificing retrieval quality. 
\sysname is designed to meet this challenge by adopting the Dynamic Wavelet Matrix (DWM), an innovative extension of the wavelet matrix—a succinct data structure renowned for its space efficiency and fast access primitives~\cite{gog2014optimized,dietzfelbinger2008succinct}—augmented to support dynamic updates for streaming agentic memory workloads.
At a high level, \sysname employs a dual-representation strategy: it stores memory content as lossless token-ID sequences for exact reconstruction (Content DWM), and parallel binary signatures for semantic search (Signature DWM). 
These two streams are co-indexed, enabling fast, bitwise retrieval directly in the compressed domain. 
The Signature DWM is constructed using random indexing, which produces compact binary representations of semantic content. 
Queries are executed via Hamming-ball search over these signatures, allowing fast, approximate matching with minimal computational cost. 
This design preserves both the fidelity of raw content and the semantic richness required for accurate, context-aware recall, while ensuring scalability and responsiveness in long-horizon agentic deployments.
Our main contributions are:
\begin{enumerate}[topsep=0pt,itemsep=-1ex,partopsep=1ex,parsep=1ex, leftmargin=*]
\item \textbf{Fundamentally New Memory Substrate.} We introduce a token-free memory substrate that replaces dense vector representations with binary signatures and token-ID streams, enabling a shift away from embedding-heavy designs toward lightweight structures.
\item \textbf{\sysname Module.} A contextual memory system built on the Dynamic Wavelet Matrix (DWM), supporting streaming writes and co-indexing of semantic and exact representations. This design enables ultra-fast Hamming-ball search in a compressed domain, while achieving high compression and superior retrieval latency and storage efficiency.
\item \textbf{Experimental Validation.} Extensive experiments demonstrate that \sysname achieves up to 31$\times$ faster end-to-end retrieval and 14$\times$ lower per-query token cost, while preserving task accuracy across both LoCoMo and LongMemEval benchmarks.
\end{enumerate}


%% file: 2_back-motiv_v1.1.tex
\section{Background and Motivation}\label{sec: background}
\begin{figure}[!t]
\centering
  \includegraphics[width=0.9\columnwidth]{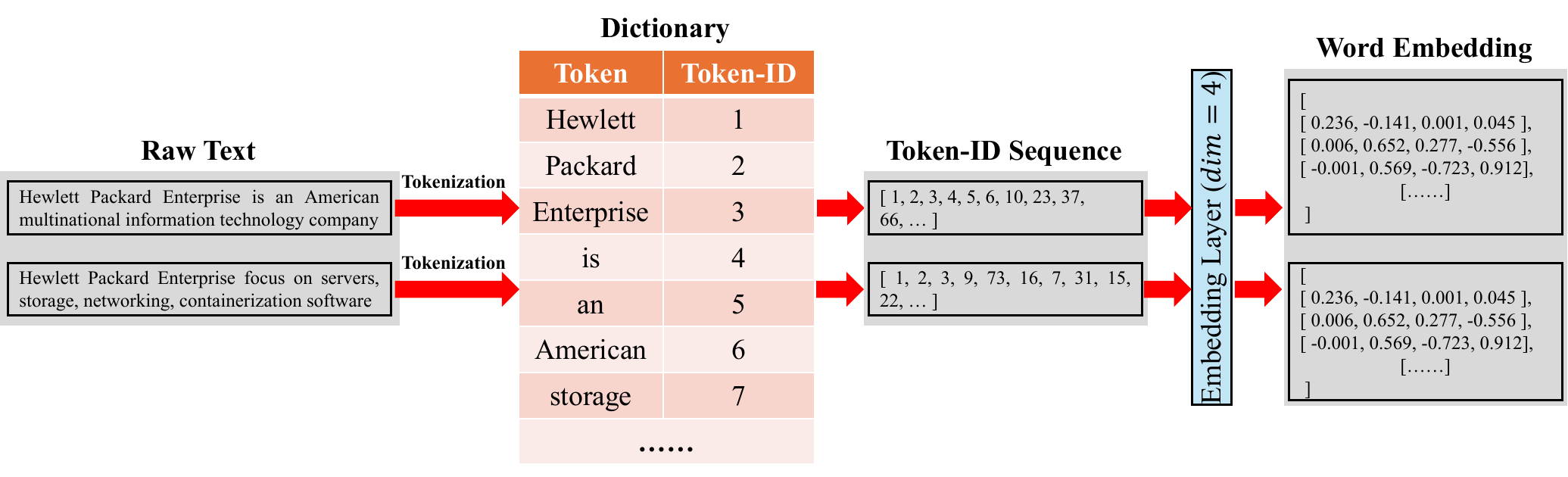}
  \caption{Illustration of raw text, token-id, and word embedding.}
  \label{fig:token_id}
  \vspace{-12pt}
\end{figure}
\subsection{Memory Representation and Management}

Agentic AI systems rely on a memory system to persist information across iterative reasoning cycles. 
Existing agentic memory systems typically represent memory content using either high-dimensional embeddings or structured graphs. 
In Retrieval-Augmented Generation (RAG)~\cite{zhang2024interactive,kagaya2024rap,singh2025agentic}, raw text is embedded into dense vectors and stored in vector databases, enabling semantic similarity search. 
Knowledge Graph (KG)-based systems~\cite{rasmussen2025zep,xu2025mem} encode memory as entity–relation–entity triples, supporting multi-hop traversal and schema-aware reasoning in graph databases such as Neo4j~\cite{guia2017graph}. 
Hybrid systems like A-Mem~\cite{xu2025mem} combine both approaches to balance semantic richness and structural precision.

While these representations offer expressive retrieval capabilities, they introduce significant performance bottlenecks. 
Embedding-based systems suffer from high computational cost during both ingestion and retrieval: memory insertion requires expensive embedding generation and preprocessing (e.g., summarization), while retrieval relies on costly vector similarity computations~\cite{mei2024aios,arora2020contextual}. 
Graph-based systems, though more interpretable, incur latency from multi-hop traversal and schema resolution. 
These inefficiencies are particularly problematic in agentic workflows, where memory is frequently updated and queried across observe–plan–act–learn cycles. 
Slow memory operations can stall agent execution, reduce throughput, and degrade responsiveness.

To support scalable, high-performance agents, memory systems must enable fast, streaming writes and low-latency recall—without sacrificing retrieval quality. 
This motivates our exploration of compression-native representations and efficient indexing mechanisms that can meet the demands of long-horizon, multi-iteration agentic deployments.

\subsection{Performance Analysis}\label{sec: performance}
\begin{figure}[!t]
\centering
  \includegraphics[width=0.95\columnwidth]{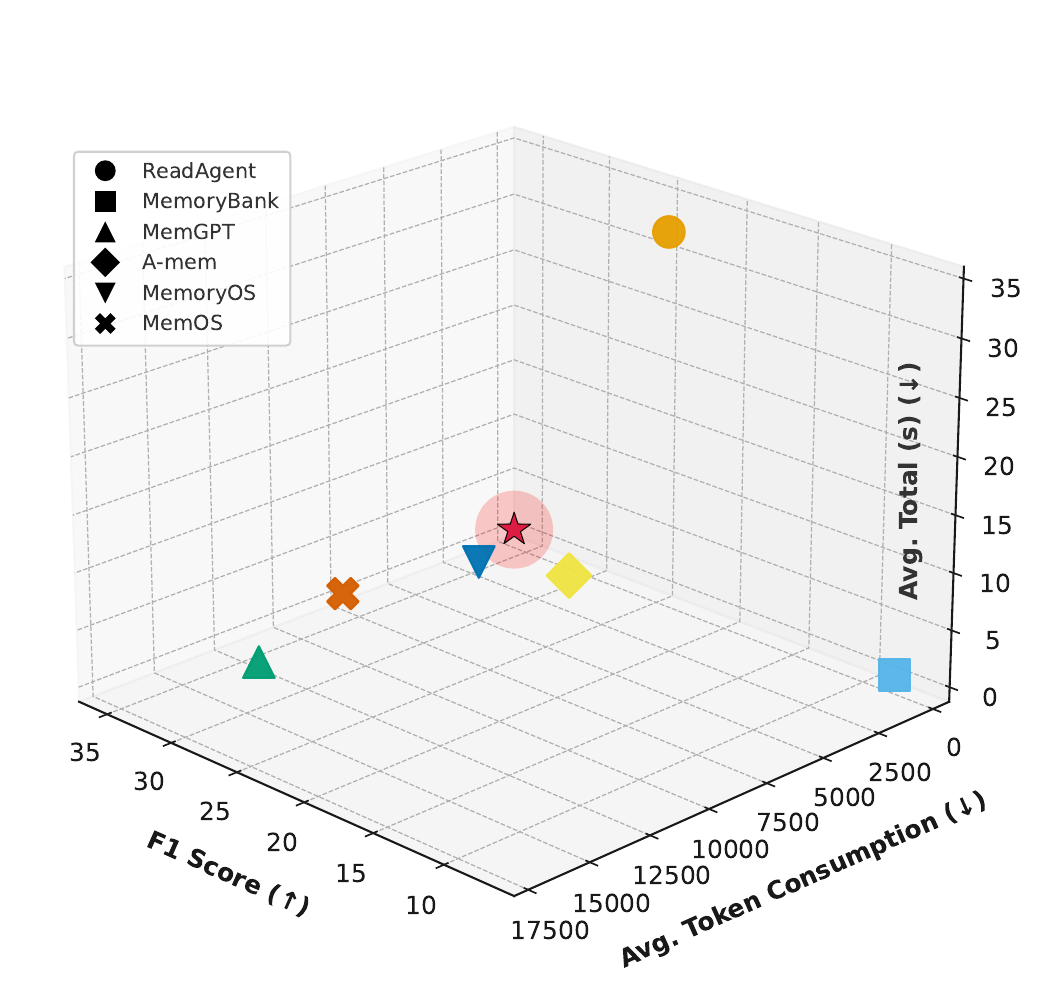}
  \caption{An analysis of SOTA agent memory systems across three critical metrics. Lower values are better for Avg. Token Consumption and Avg. Total Time, while higher is better for F1 Score. The plot illustrates that existing systems force a compromise, as none are able to simultaneously achieve high accuracy and high efficiency in the ideal design space indicated by the red star marker.}
  \label{fig:trade_off}
  \vspace{-12pt}
\end{figure}
\begin{figure}[!t]
\centering
  \includegraphics[width=0.95\columnwidth]{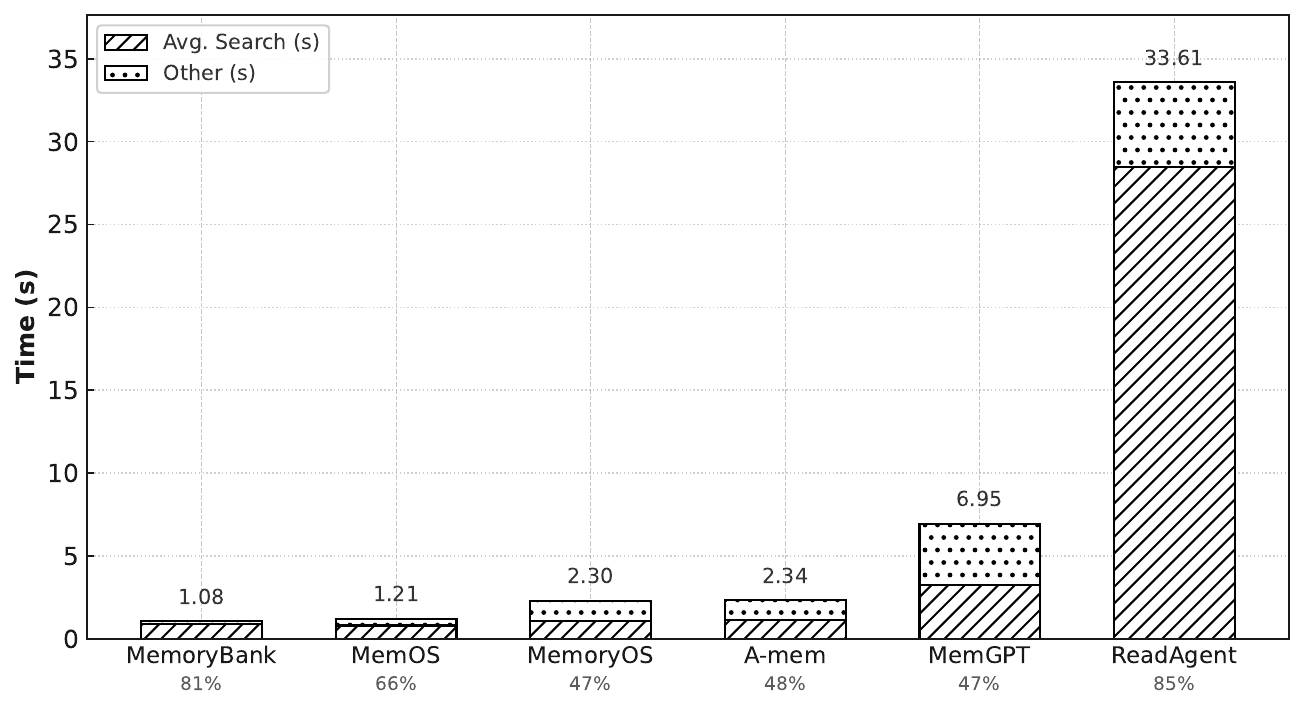}
  \caption{Breakdown of the end-to-end retrieval latency for SOTA agentic AI memory modules.}
  \label{fig:search_rate}
  \vspace{-12pt}
\end{figure}

To understand the design trade-offs in existing agentic memory systems, we evaluated six representative state-of-the-art (SOTA) modules—ReadAgent~\cite{lee2024human}, MemoryBank~\cite{zhong2024memorybank}, MemGPT~\cite{packer2023memgpt}, A-Mem~\cite{xu2025mem}, MemoryOS~\cite{kang2025memory}, and MemOS~\cite{li2025memos}—across three critical metrics: retrieval accuracy (F1 score), operational cost (average token consumption per query), and user-perceived latency (average total query time). The evaluation was conducted on the LoCoMo benchmark~\cite{maharana2024evaluating}, which simulates long-horizon agentic tasks with frequent memory interactions.

As shown in Figure~\ref{fig:trade_off}, current memory systems force developers into a difficult compromise. High-accuracy designs like MemGPT and A-Mem achieve strong F1 scores but incur significant latency and token overhead due to embedding generation, summarization, and multi-stage retrieval. Conversely, lightweight systems such as MemoryBank reduce latency and cost but suffer from degraded recall quality. None of the evaluated systems simultaneously optimize all three axes, leaving the ideal region of the design space—high accuracy with low latency and cost—unoccupied.

While above analysis focuses on retrieval efficiency, the situation is further aggravated by insertion-side overhead during memory growth. In RAG, these arise from chunking, embedding, and index updates~\cite{zhong2024mix}; in KG, from fact insertions and graph index maintenance~\cite{anadiotis2024dynamic,wandji2024improving}; and in hybrid memories (e.g., A-Mem), from the additional note creation and cross-linking steps that improve read quality but inflate token and amortization cost. These write-path penalties add latency even before retrieval begins, exacerbating the trade-off illustrated in Figure~\ref{fig:trade_off}~\cite{xu2025mem}.

To pinpoint the root cause of this inefficiency, Figure~\ref{fig:search_rate} shows a breakdown of end-to-end retrieval latency. 
The results show that the memory search phase dominates total execution time across all architectures. 
In ReadAgent, vector similarity search accounts for 85\% of recall latency. 
Even in more streamlined systems like MemoryBank, search operations consume 81\% of the time. 
Hybrid systems such as A-Mem and MemoryOS, which incorporate structured memory layouts and multi-hop reasoning, spend nearly half of their runtime in retrieval (48\% and 47\%, respectively).

These findings highlight a fundamental limitation: the retrieval substrate itself—whether based on dense vectors or graph traversal—is the primary bottleneck. In agentic workflows, where memory is accessed and updated repeatedly across observe–plan–act–learn cycles, such inefficiencies compound rapidly. Slow retrieval stalls the planning, while costly ingestion limits memory growth. To support scalable, responsive agents, we need a memory system that rethinks the underlying data structures and representations, enabling fast, streaming writes and low-latency recall at scale.

\subsection{Tokenized Memory and Succinct Data Structures}
Given that LLMs natively operate on integer sequences called token-IDs~\cite{qu2024tokenrec,yu2024breaking}, we adopt token-IDs as the fundamental representation of memory. This compact, model-native format avoids repeated and costly tokenization cycles, enabling efficient storage and manipulation. More importantly, representing memory as integer sequences allows us to leverage powerful succinct data structures—such as the Wavelet Matrix~\cite{gog2014optimized,claude2012wavelet,dietzfelbinger2008succinct}—to build a high-performance retrieval system that operates directly in the compressed domain.\\
\textbf{Wavelet Matrix.}
Succinct data structures are compact representations that approach the information-theoretic minimum space while supporting fast queries directly on compressed data~\cite{dietzfelbinger2008succinct,shamir2006universal}. 
Among these, the Wavelet Matrix~\cite{gog2014optimized,claude2012wavelet} is particularly well-suited for representing long sequences of discrete symbols, such as token-IDs in LLMs.
It arranges the bits of each symbol into a multi-level structure and supports three core operations with logarithmic time complexity:
\begin{itemize}[topsep=0pt,itemsep=-1ex,partopsep=1ex,parsep=1ex, leftmargin=*]
  \item $access(i)$: Retrieve the symbol at position $i$.
  \item $rank(c, i)$: Number of symbol $c$ appears in prefix $[0, i)$.
  \item $select(c, j)$: Position of the $j$-th occurrence of symbol $c$.
\end{itemize}

However, canonical wavelet matrices are static and operate over a single homogeneous sequence, making them incompatible with agentic workloads where memory is continuously appended and must be immediately available for retrieval. 
Rebuilding the entire matrix for each new memory entry would be computationally prohibitive~\cite{claude2012wavelet}, especially in long-horizon deployments.
Moreover, as detailed in Section~\ref{sec: dwm}, \sysname introduces two distinct but interrelated data streams, i.e., memory content and memory signatures, that must be co-indexed to support efficient retrieval. 
Nevertheless, the standard wavelet matrix~\cite{gog2014optimized,claude2012wavelet} lacks native support for co-indexing heterogeneous sequences, limiting its applicability in our design.\\
\textbf{Semantic Hashing via Random Indexing.} To enable efficient semantic search, we leverage Semantic Hashing, a form of Locality-Sensitive Hashing (LSH)~\cite{indyk1998approximate}, to convert high-dimensional vectors into compact binary signatures. 
Using a computationally inexpensive method called Random Indexing~\cite{kanerva2000random}, we project each vector against a set of random hyperplanes to generate its signature. This ensures that semantically similar vectors are mapped to signatures with a small Hamming distance (i.e., differing in only a few bits)~\cite{norouzi2012hamming,labib2019hamming}. This crucial property allows us to replace the expensive k-Nearest Neighbor (k-NN) search~\cite{de2002efficient} over floating-point vectors with an ultra-fast search for neighbors within a small Hamming radius, an operation that can be massively accelerated using native bitwise CPU operations~\cite{seshadri2016buddy}.

\begin{figure*}[!t]
\centering
  \includegraphics[width=0.95\linewidth]{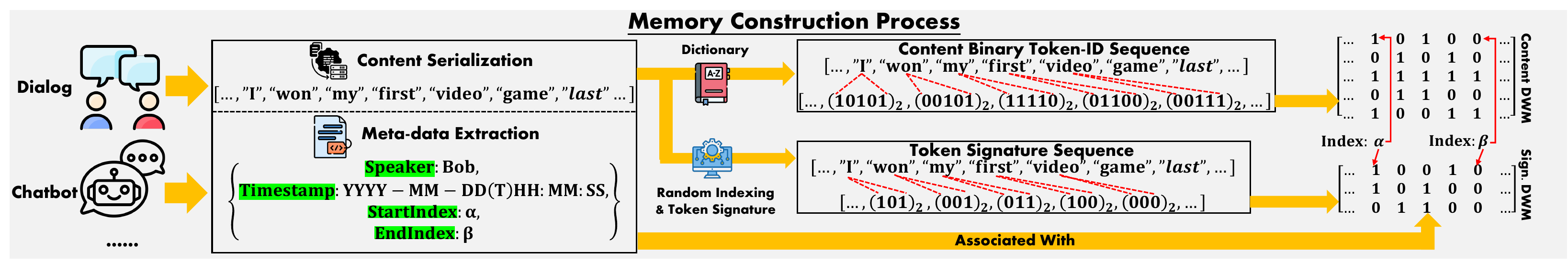}
  \caption{Illustration of the memory construction pipeline in Hippocampus. \textbf{DWM} denotes our proposed \textbf{D}ynamic \textbf{W}avelet \textbf{M}atrix, while the subscript $(\cdot)_2$ indicates the binary representation of an integer, for example, token-id. The first row of the DWM serves as the entry-level index, marking the start and end positions of each token in the \textbf{Content Serialization} (e.g., \textbf{StartIndex} $\alpha$ and \textbf{EndIndex} $\beta$).}
  \label{fig:overall_construction}
\end{figure*}
\begin{figure*}[!t]
\centering
  \includegraphics[width=0.95\linewidth]{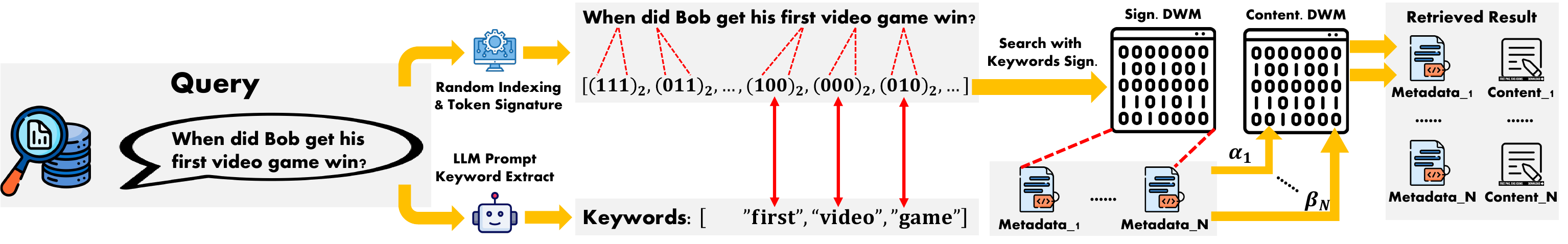}
  \caption{Illustration of the memory query pipeline in the Hippocampus. An LLM first extracts keywords from the natural language query. These keywords are converted into binary signatures and used to perform a fast, approximate search on the Signature (Sign.) Dynamic Wavelet Matrix (DWM), identifying candidate metadata blocks. The indices (e.g., $\alpha_1$) from the retrieved metadata are then used to look up and reconstruct the exact, full-resolution content from the Content DWM.}
  \label{fig:overall_query}
  \vspace{-12pt}
\end{figure*}